\documentclass{article}

\usepackage{microtype}
\usepackage{graphicx}
\usepackage{subcaption}
\usepackage{booktabs} 
\usepackage{stfloats}
\usepackage{hyperref}

\usepackage[accepted]{icml2026}

\usepackage{amsmath}
\usepackage{amssymb}
\usepackage{mathtools}
\usepackage{amsthm}
\usepackage{multirow}
\usepackage{booktabs}
\usepackage[table]{xcolor} 
\usepackage[capitalize,noabbrev]{cleveref}
\usepackage{enumitem}
\theoremstyle{plain}
\newtheorem{theorem}{Theorem}[section]

\theoremstyle{definition}
\newtheorem{definition}[theorem]{Definition}

\theoremstyle{remark}

\usepackage[textsize=tiny]{todonotes}

\icmltitlerunning{Token Pruning for In-Context Generation in Diffusion Transformers}
\newcommand{\toolName}{ToPi}

\begin{document}

\twocolumn[
  \icmltitle{\textit{To}ken \textit{P}runing for \textit{In}-Context Generation in Diffusion Transformers}




  \begin{icmlauthorlist}
    \icmlauthor{Junqing Lin}{ustc,comp}
    \icmlauthor{Xingyu Zheng}{comp,bhu}
    \icmlauthor{Pei Cheng}{comp}
    \icmlauthor{Bin Fu}{comp}
    \icmlauthor{Jingwei Sun}{ustc}
    \icmlauthor{Guangzhong Sun}{ustc}
  \end{icmlauthorlist}

  \icmlaffiliation{ustc}{University of Science and Technology of China, Anhui, China}
  \icmlaffiliation{bhu}{Beihang University, Beijing, China}
  \icmlaffiliation{comp}{Tencent PCG, China}

  \icmlcorrespondingauthor{Jingwei Sun}{sunjw@ustc.edu.cn}
  \icmlcorrespondingauthor{Guangzhong Sun}{gzsun@ustc.edu.cn}

  \icmlkeywords{Machine Learning, ICML}

  \vskip 0.3in
]



\printAffiliationsAndNotice{}  

\begin{abstract}

In-context generation significantly enhances Diffusion Transformers (DiTs) by enabling controllable image-to-image generation through reference examples. However, the resulting input concatenation drastically increases sequence length, creating a substantial computational bottleneck. Existing token reduction techniques, primarily tailored for text-to-image synthesis, fall short in this paradigm as they apply uniform reduction strategies, overlooking the inherent role asymmetry between reference contexts and target latents across spatial, temporal, and functional dimensions. 
To bridge this gap, we introduce ToPi, a training-free token pruning framework tailored for in-context generation in DiTs.
Specifically, ToPi utilizes offline calibration-driven sensitivity analysis to identify pivotal attention layers, serving as a robust proxy for redundancy estimation. Leveraging these layers, we derive a novel influence metric to quantify the contribution of each context token for selective pruning, coupled with a temporal update strategy that adapts to the evolving diffusion trajectory. Empirical evaluations demonstrate that ToPi can achieve over 30\% speedup in inference while maintaining structural fidelity and visual consistency across complex image generation tasks.

\end{abstract}

\section{Introduction}

Diffusion Transformers (DiTs)~\cite{peebles2023scalable,xing2024survey,shen2025efficient} have emerged as the dominant architecture for generative modeling by treating diverse modalities as sequences of latent tokens, effectively capitalizing on the scalability of transformers. 
Building on this foundation, in-context generation further enhances generation controllability for scenarios such as image-to-image and image-to-video by directly incorporating reference examples into the input sequence without fine-tuning~\cite{liu2024zero,wan2025,kong2024hunyuanvideo,labs2025flux1kontextflowmatching,wu2025qwenimagetechnicalreport}.
However, this paradigm extends the input sequence with reference examples, significantly increasing the computational overhead and memory footprint due to the massive number of tokens.
Furthermore, the quadratic complexity of self-attention with respect to sequence length exacerbates this issue, creating a scalability barrier for high-resolution or multi-reference generation tasks.

To mitigate the computational overhead of DiTs, prior works have primarily focused on three strategic directions: accelerating sampling schedules~\cite{lu2022dpm,song2023consistency,salimans2022progressive,huang2025self}, compressing model parameters~\cite{chen2024QDiT,li2024svdquant,wang2025sparsedm,zhao2024dynamic}, and dynamic computation reduction~\cite{liu2025reusing,zhang2025spargeattn,bolya2022tome}. Within the third direction, token reduction  methods~\cite{lu2025toma,bolya2023tomesd,fang2025attend,kim2024token} are especially attractive: by explicitly limiting the number of active tokens, they directly attack the primary bottleneck induced by long sequences.
However, existing token reduction strategies prove ineffective for in-context generation. These methods indiscriminately process reference context and target latents, and overlook that context utility depends on its relevance to the current denoising state. Consequently, existing methods either retain excessive redundancy or blindly prune critical semantic anchors, degrading generation fidelity.

In this paper, we seek a principled answer to a central question in efficient in-context DiTs: when and which tokens can be removed without compromising fidelity? 
Our starting point is the observation that reference context tokens and target latent tokens play asymmetric roles during denoising. 
Our systematic analysis reveals that this asymmetry manifests across three interconnected dimensions. 
(1) Spatially, semantic influence concentrates within a small subset of pivotal layers rather than distributing uniformly.
(2) Temporally, redundancy rises as target latents progressively assimilate reference context.
(3) Functionally, more complex tasks require denser semantic anchors than simple edits.

Motivated by these findings, we introduce {\toolName}, 
a training-free token pruning framework tailored for in-context generation in DiTs.
Specifically, {\toolName} employs three mechanisms to handle the inherent asymmetry of diffusion tasks. First, it employs a calibration-driven sensitivity analysis to pinpoint pivotal attention layers, serving as a robust proxy for redundancy estimation.
Leveraging these layers, we derive a novel influence metric that quantifies the contribution of each reference token, allowing us to selectively discard non-essential context while preserving semantic anchors. Finally, {\toolName} periodically updates the token selection to synchronize with the evolving diffusion trajectory. Empirical evaluations on Flux.1-Kontext~\cite{labs2025flux1kontextflowmatching} and Qwen-Image-Edit~\cite{wu2025qwenimagetechnicalreport} demonstrate that {\toolName} can achieve over 30\% speedup in inference on the AnyEdit~\cite{yu2025anyedit} benchmark  without compromising structural fidelity or visual consistency.

This paper makes the following contributions:
\begin{itemize}
    \item We systematically analyze token redundancy during in-context generation in DiTs. We demonstrate that context utility varies across spatial, temporal, and functional dimensions. This characterization highlights the role asymmetry between reference context and target latents, exposing the inadequacy of uniform pruning strategies.
    \item We introduce {\toolName}, a training-free framework grounded in signal contribution preservation maximization. We formulate a novel influence metric that identifies importance tokens, coupled with a periodic update strategy to align with the temporal dynamics of diffusion.
    \item We perform extensive empirical evaluations across diverse image generation benchmarks. The results demonstrate that our approach achieves significant reductions in inference latency without compromising the visual quality or semantic coherence of the output.
\end{itemize}

\section{Related Work}

\textbf{Diffusion Model Acceleration.}
Research on accelerating Diffusion Probabilistic Models (DPMs)~\cite{ho2020denoising,song2020denoising,dhariwal2021diffusion} primarily addresses three distinct axes of inefficiency. 
The first concerns extensive sampling iterations, where advanced solvers~\cite{lu2022dpm, song2020denoising} or distillation techniques~\cite{salimans2022progressive, song2023consistency,huang2025self} compress the inference trajectory. 
The second targets parameter redundancy, employing quantization~\cite{wu2024ptq4dit,chen2024QDiT, li2024svdquant} or pruning~\cite{wang2025sparsedm, zhao2024dynamic} to compact model weights. 
Our work operates on the third axis: dynamic computation reduction, which optimizes data flow at run-time. 
It includes temporal strategies~\cite{ma2023deepcache, zou2024accelerating, lou2024token, liu2025reusing} that cache or reuse features across timesteps,
as well as spatial strategies that eliminate calculation amount via sparse attention~\cite{zhang2025fast,yang2025sparse,zhang2025spargeattn,zhang2025vsa} or through \textit{token reduction} methods~\cite{bolya2023tomesd,fang2025attend, kim2024token, lu2025toma} that explicitly reduce sequence length.

\textbf{Token Reduction: From Discriminative to Generative.}
Token reduction was initially developed for discriminative Vision Transformers (ViTs)~\cite{rao2021dynamicvit,zhai2022lit} and Multimodal LLMs~\cite{wu2023multimodal,caffagni2024revolution}, employing pruning~\cite{rao2021dynamicvit, chen2024image,zhang2024fastervlm,zhangsparsevlm,zhang2025beyond, zou2025don} or merging~\cite{bolya2022tome, shang2025llava,li2025mergevq} to alleviate computational overhead.
However, applying existing reduction methods to visual in-context generation encounters two structural mismatches. 
First, there is a \textit{purpose mismatch}: discriminative objectives prioritize high-level semantic abstraction, whereas generative modeling requires dense signal propagation to maintain spatial fidelity. 
Second, there is a \textit{contextual mismatch} in existing DPM adaptations~\cite{bolya2023tomesd,kim2024token,smith2024todo,wu2025importance,fang2025attend}. Most of them are optimized for independent image generation and fail to distinguish the distinct roles of reference context and target latents, neglecting the critical interactions required for in-context generation.
{\toolName} bridges these mismatches by introducing a generative-oriented influence metric to preserve reconstruction features, while simultaneously addressing the redundancy specific to reference tokens to enable efficient context-aware generation.

\section{Preliminaries and Motivation}
\subsection{Preliminaries}
\label{sec:preliminaries}
We focus on conditional diffusion models operating within the latent space of a pre-trained VAE. The model functions by iteratively denoising noisy tokens $\mathcal{G}^t  \in \mathbb{R}^{|\mathcal{G}^t| \times d}$ across timesteps $t=T, \dots, 1$ to recover the clean state $\mathcal{G}^0$. Within this framework, in-context generation facilitates image generation conditioned on a fixed sequence of reference tokens $\mathcal{C} \in \mathbb{R}^{|\mathcal{C}| \times d}$ to improve the controllability of the synthesis process. Fundamentally, in-context generation operates by concatenating these reference tokens with the noisy tokens $\mathcal{G}$ along the sequence dimension.
This yields a unified input sequence $\mathbf{Z}^t \in \mathbb{R}^{(|\mathcal{G}^t| + |\mathcal{C}|) \times d}$:
\begin{equation}
\mathbf{Z}^t = [\mathcal{G}^t; \mathcal{C}],
\end{equation}
where $[\cdot; \cdot]$ denotes the concatenation operation. Then, the self-attention mechanism within each DiT block operates globally on this joint sequence. 
Let $\mathbf{H}$ represent the intermediate features at a given layer; the query, key, and value projections are defined as $\mathbf{Q}=\mathbf{H}\mathbf{W}_Q$, $\mathbf{K}=\mathbf{H}\mathbf{W}_K$, and $\mathbf{V}=\mathbf{H}\mathbf{W}_V$. The resulting attention matrix captures full bidirectional dependencies:
\begin{equation}
\mathbf{A} = \mathrm{softmax}\left( \frac{\mathbf{Q}\mathbf{K}^\top}{\sqrt{d}} \right).
\end{equation}
While this unified formulation ensures dense information flow between the target latents and reference contexts, it incurs a computational cost that scales quadratically with the total sequence length.

\subsection{Motivation: Redundancy in Contextual Attention}
\label{sec:motivation}

We hypothesize that current architectures suffer from significant inefficiency by failing to account for the role asymmetry between reference context and target latents. Rather than requiring a dense, uniform integration, the generative process relies on sparse, discriminative visual cues.
To quantify this, we define the target-aware attention score $S_j$ for a specific reference token $j \in \mathcal{C}$. This metric represents the average attention weight that token $j$ receives from all noisy tokens $\mathcal{G}$, averaged across heads $\mathcal{H}$:
\begin{equation}
    S_j = \frac{1}{|\mathcal{G}| \cdot |\mathcal{H}|} \sum_{h \in \mathcal{H}} \sum_{i \in \mathcal{G}} A_{ij, h}.
    \label{eq:analysis_metric}
\end{equation}
Our empirical analysis confirms that token importance is highly non-uniform, characterized by three dimensions of asymmetry: spatial concentration, temporal evolution, and functional anchoring.

\begin{figure}[b]
    \centering
    \begin{subfigure}{0.495\linewidth}
        \centering
        \includegraphics[width=1\linewidth]{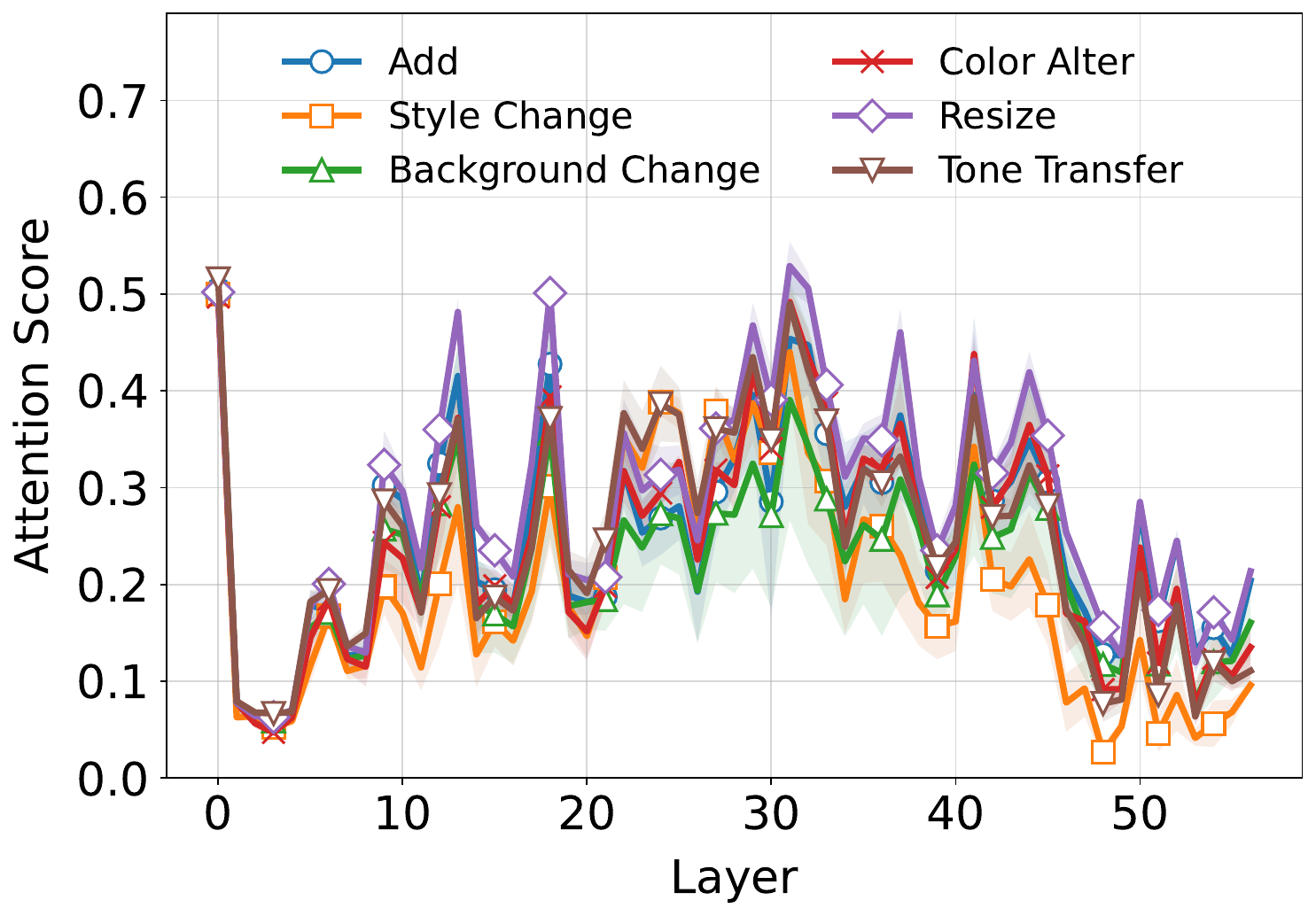}
        \caption{Flux.1-Kontext}
    \end{subfigure}
    \begin{subfigure}{0.495\linewidth}
        \centering
        \includegraphics[width=1\linewidth]{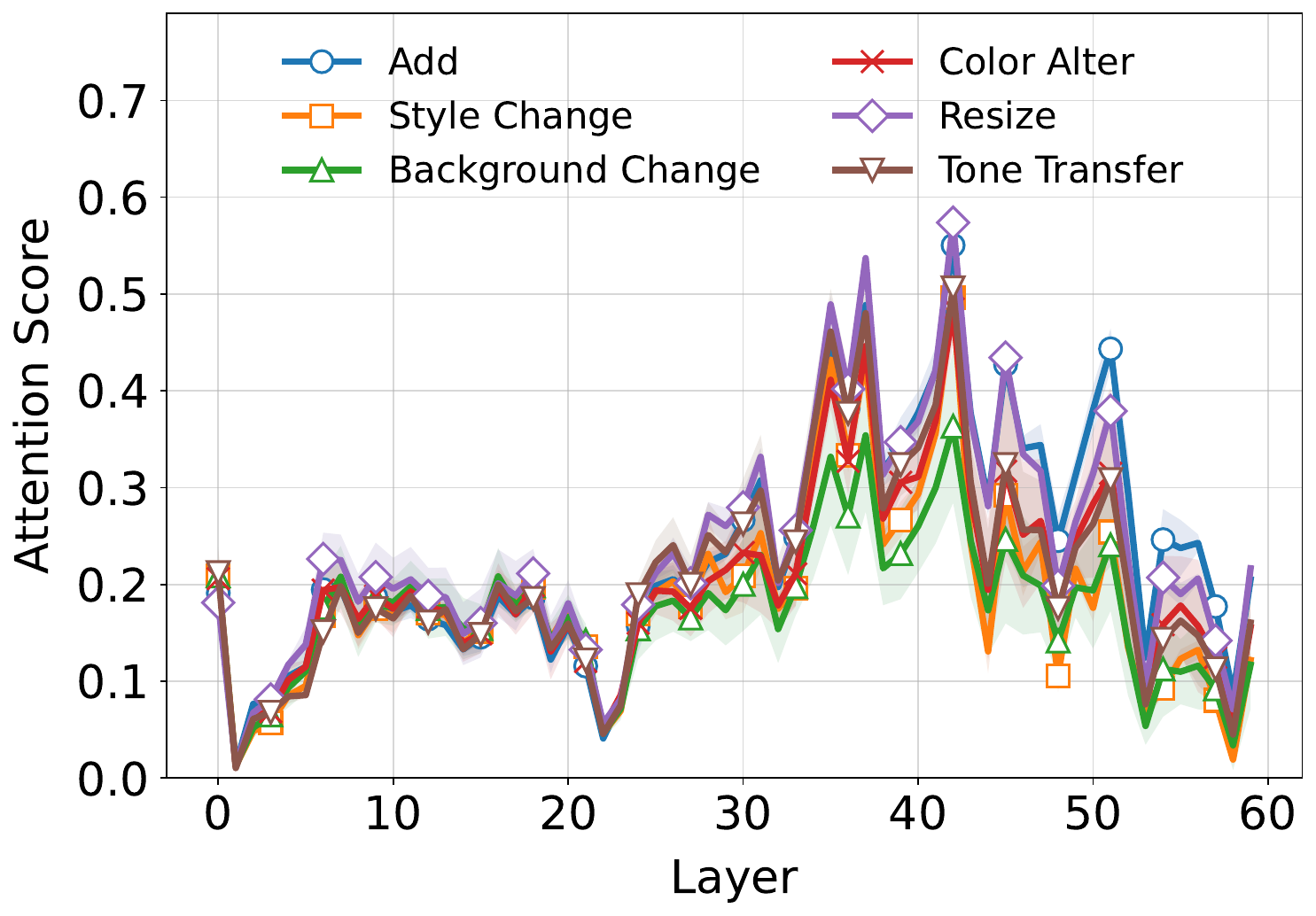}
        \caption{Qwen-Image-Edit}
    \end{subfigure}
    \caption{Layer-wise distribution of the total target-aware attention score ($\sum S_j$), averaged across timesteps.}
    \label{fig:layer_attn}
\end{figure}

\paragraph{Spatial: Concentration in Pivotal Layers.}
Fig.~\ref{fig:layer_attn} reveals spatial asymmetry in attention distribution across network layers. Rather than exhibiting uniform semantic diffusion, attention mass concentrates heavily within a sparse subset of pivotal layers while the remaining layers contribute negligible weight. Meanwhile, this stratification is input-invariant, indicating that context retrieval is a layer-specialized process rather than a global property. Consequently, these pivotal layers function as sufficient proxies for global token importance, enabling reliable redundancy estimation without requiring full-layer evaluation.

\begin{figure}[t]
    \centering
    \includegraphics[width=0.8\linewidth]{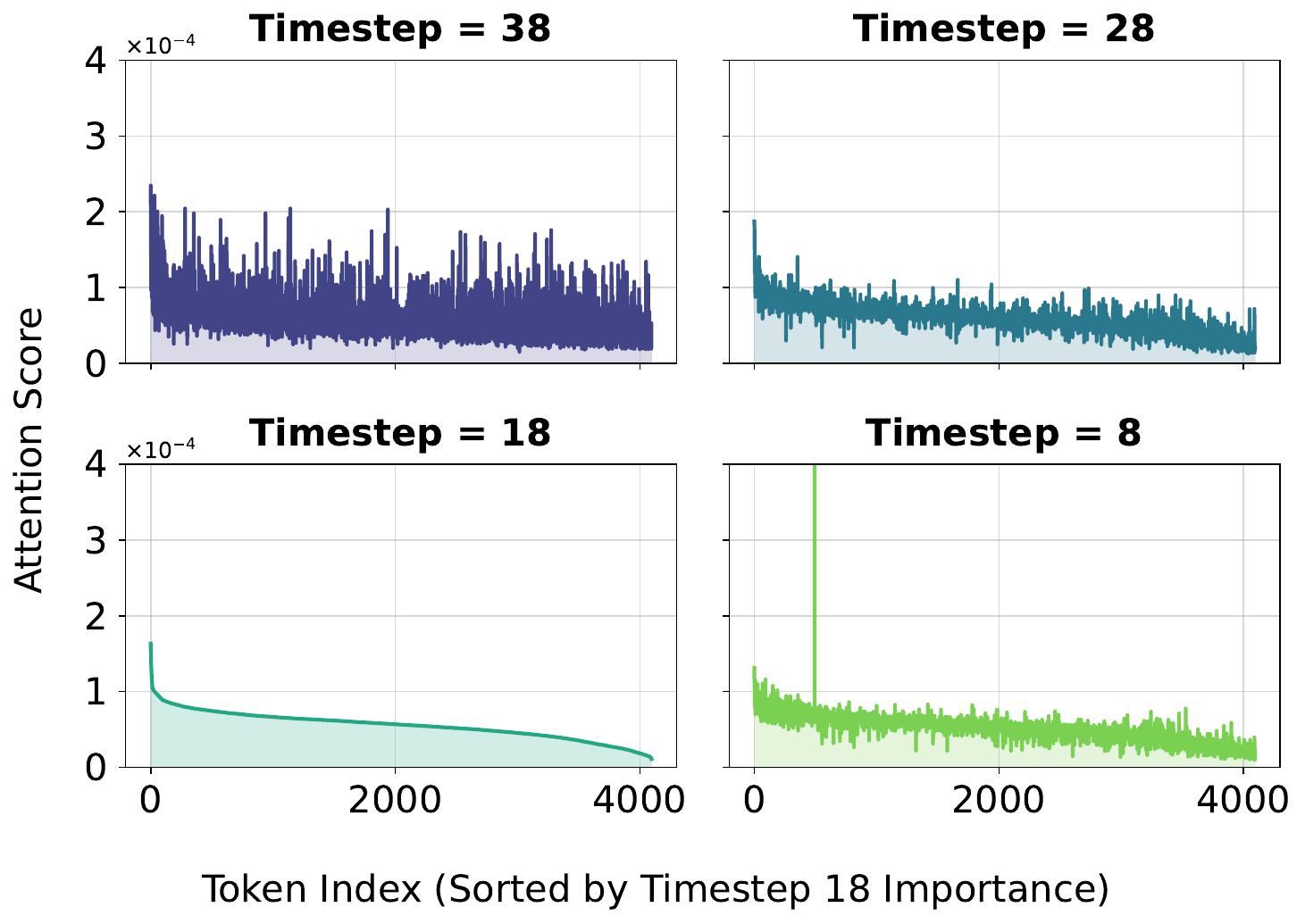}
    \caption{Evolution of individual target-aware attention scores ($S_j$) for reference tokens across timesteps, averaged across layers.}
    \label{fig:token_attn}
\end{figure}

\begin{figure}[ht]
    \centering
    \begin{subfigure}{0.495\linewidth}
        \centering
        \includegraphics[width=1\linewidth]{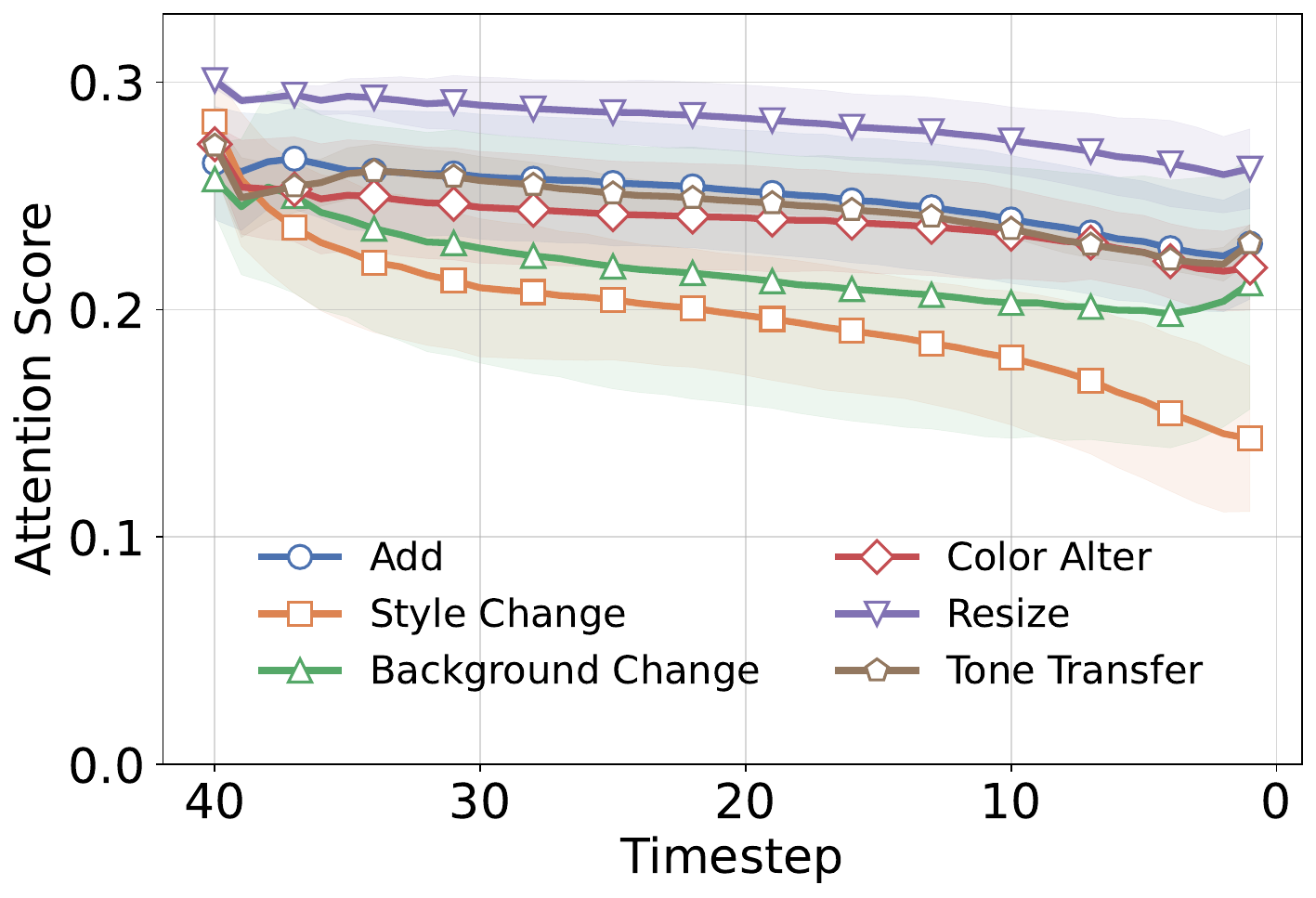}
        \caption{Flux.1-Kontext}
    \end{subfigure}
    \begin{subfigure}{0.495\linewidth}
        \centering
        \includegraphics[width=1\linewidth]{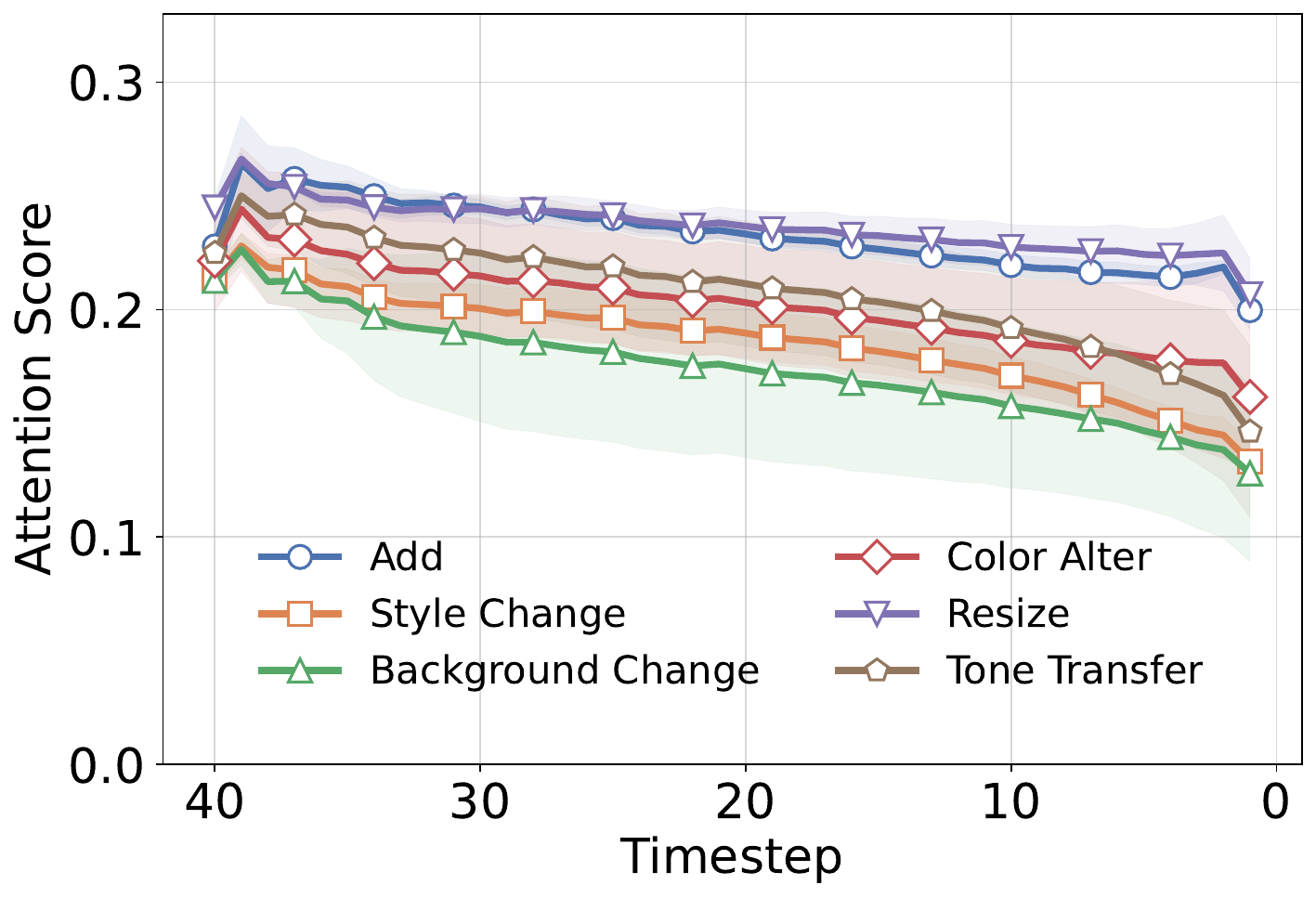}
        \caption{Qwen-Image-Edit}
    \end{subfigure}
    \caption{Temporal trajectory of the total attention score ($\sum S_j$) under different task conditions, averaged across layers.}
    \label{fig:step_attn}
\end{figure}

\paragraph{Temporal: Evolution over Timesteps.} 
Contextual interaction exhibits two temporal dynamics. \textit{(1) Fluctuating Token Relevance.} Fig.~\ref{fig:token_attn} reveals that the relative importance of individual tokens shifts dynamically throughout the generation process. Consequently, strategies relying on a fixed set of selected tokens during the entire denoising process are inherently suboptimal. \textit{(2) Global Attentional Decay.} As shown in Fig.~\ref{fig:step_attn}, the total score follows a trajectory of temporal decay. Early high-noise stages rely heavily on reference features for structural guidance, whereas later stages transition toward self-contained refinement.

\paragraph{Functional: Task-Dependent Anchoring.}

Quantitatively, Fig.~\ref{fig:step_attn} reveals that total reference attention remains consistently low ($< 0.3$), indicating that the generation process is predominantly driven by self-attention within the noisy tokens. This data characterizes reference tokens functioning primarily as auxiliary semantic anchors rather than dense constructive elements. Crucially, the reliance on these anchors is dictated by functional asymmetry: high-fidelity tasks (e.g., Resize) sustain elevated attention levels to preserve structural details, whereas loose-constraint tasks (e.g., Style Change) exhibit a rapid decline in attention mass. This task-dependent variability mandates that computational allocation be dynamically modulated to match the specific functional demands of the intervention.

\begin{figure*}[t]
    \centering
    \includegraphics[width=1.0\linewidth]{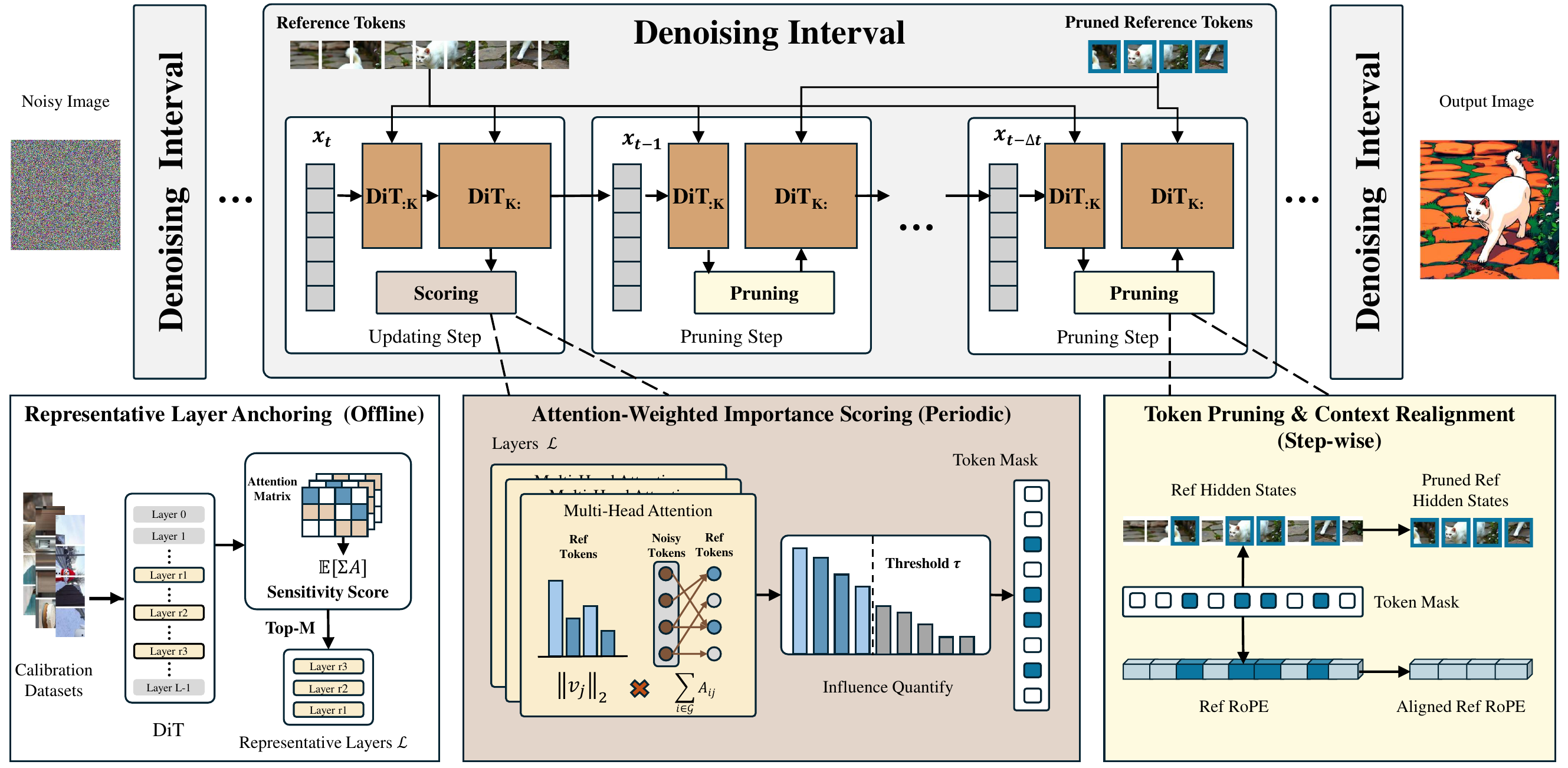}
\caption{
    \textbf{Overview of \toolName.} 
    The upper panel depicts the denoising trajectory, which contains multiple denoising intervals. Each interval comprises a mask updating step and multiple token pruning steps. This process is controlled by three stratified mechanisms detailed in the lower panels:
    \textbf{(1) Offline Calibration:} One-time identification of representative layers $\mathcal{L}$ of a DiT to filter out insensitive layers.
    \textbf{(2) Periodic Scoring:} Importance scoring and mask updates every $\Delta T$ steps to adapt to semantic shifts.
    \textbf{(3) Step-wise Pruning:} Instant token pruning and context realignment applied at each denoising step.
}

    \label{fig:overview}
\end{figure*}

\section{Method}
\label{sec:method}
As analyzed in Sec.~\ref{sec:motivation}, the reference context acts as a supportive resource, from which target latents selectively attend to only sparse semantic regions. 
To leverage this inherent redundancy, we introduce \textbf{\toolName}, a training-free and dynamic token pruning framework tailored for in-context generation in DiTs. 
Rather than processing the full sequence, {\toolName} dynamically extracts a minimal sufficient subset $\mathcal{C}'$ from the reference tokens $\mathcal{C}$, retaining only the semantic core essential for high-fidelity generation. 

As illustrated in Fig.~\ref{fig:overview}, {\toolName} employs three mechanisms sequentially. (1) \textbf{Representative Layer Anchoring} identifies semantically dense layers offline; (2) \textbf{Attention-Weighted Importance Scoring} periodically computes token importance scores and updates the retention mask during inference; 
and (3) \textbf{Token Pruning \& Context Realignment} applies the mask at each timestep, directly accelerating the generative process.
The complete algorithmic formulation is provided in Appendix~\ref{app:algo}.

\subsection{Representative Layer Anchoring}
\label{sec:layer_selection}

Modern diffusion architectures typically consist of numerous attention layers. Using all layers to estimate importance imposes two primary constraints. 
First, aggregating attention statistics across the entire network introduces substantial overhead, which can negate the efficiency gains from pruning.
Second, empirical analysis reveals that layers exhibit significant heterogeneity in their sensitivity to the context $\mathcal{C}$. 
The inclusion of layers with negligible sensitivity introduces noise, thereby compromising the fidelity of token importance rankings. To mitigate these limitations, we isolate a subset of \textit{Representative Layers}. These layers encapsulate the dominant interactions with the visual context and serve as a robust proxy for the global context processing of the full model.

Guided by our observation in Sec.~\ref{sec:motivation}, which demonstrates that the spatial location of pivotal layers remains consistent across diverse tasks,  we constructed $\mathcal{D}_{calib}$ to be a stratified subset covering multiple categories. 
Consider the $\ell$-th attention layer, which incorporates a set of attention heads $\mathcal{H}$. We denote the query features (noisy tokens) as $\mathbf{Q}^{(\ell)} \in \mathbb{R}^{|\mathcal{H}| \times |\mathcal{G}| \times d}$ and the key features (reference tokens) as $\mathbf{K}^{(\ell)} \in \mathbb{R}^{|\mathcal{H}| \times |\mathcal{C}| \times d}$. The normalized attention weights are given by $\mathbf{A}^{(\ell)} = \text{Softmax}(\frac{\mathbf{Q}^{(\ell)}(\mathbf{K}^{(\ell)})^\top}{\sqrt{d}})$. We define the \textit{Context Sensitivity Score} $S^{(\ell)}$ as the expected attention mass accumulated across the calibration set:
\begin{equation}
    S^{(\ell)} = \mathbb{E}_{\mathbf{x} \sim \mathcal{D}_{\text{calib}}} \left[  \sum_{h\in \mathcal{H}}\sum_{i \in \mathcal{G}}\sum_{j \in \mathcal{C}} A_{hij}^{(\ell)} \right].
\end{equation}
A higher value of $S^{(\ell)}$ indicates that the layer exhibits a stronger dependency on the provided context $\mathcal{C}$, distinguishing it as a Representative Layer.
We isolate the Top-$M$ layers with the highest sensitivity scores to construct the representative set $\mathcal{L}$:
\begin{equation}
    \mathcal{L} = \operatorname{Top-M}_{\ell \in \{0,\dots,L-1\}} \left\{ S^{(\ell)} \right\}.
\end{equation}

\subsection{Attention-Weighted Importance Scoring}
\label{sec:info_quantity}

\paragraph{Theoretical Objective.}
To preserve contextual semantic integrity, we formulate the pruning objective as a signal contribution preservation problem.
Specifically, we consider the attention output $\mathbf{y}_i$ for the $i$-th token as a linear combination of value vectors from the entire input sequence $\mathcal{G} \cup \mathcal{C}$:
\begin{equation}
\mathbf{y}_i = \sum_{k \in \mathcal{G} \cup \mathcal{C}} A_{ik} \mathbf{v}_k.
\end{equation}
By decomposing this output into individual signal components, the specific contribution of the $k$-th token to the $i$-th output is captured by the vector $\mathbf{c}_{ik}$:
\begin{equation}
\mathbf{c}_{ik} \coloneqq A_{ik} \mathbf{v}_k.
\end{equation}
In this formulation, token importance is quantified by the magnitude of its contribution to the final representation, effectively integrating attention weights with feature strength. This metric ensures that a token with moderate attention weights can still exert significant influence if it possesses a large feature vector.
To minimize cumulative information loss across the target latents, we seek to identify the token $k$ yielding the minimal aggregate signal norm:
\begin{equation}\min_{k}  \sum_{i \in \mathcal{G}}\|\mathbf{c}_{ik}\|_2 = \min_{k}  \sum_{i \in \mathcal{G}} A_{ik} \|\mathbf{v}_k\|_2.
\end{equation}
\paragraph{Metric Formulation.}
Building upon this theoretical insight, we generalize this criterion to the full Transformer architecture. Given that signal contributions are additive across independent subspaces, we assess a token's global importance by aggregating its contributions. Accordingly, we define the \textbf{Token Influence Score} $I_j$ as the sum of weighted norms across all attention heads $\mathcal{H}$ and layers $\mathcal{L}$:
\begin{definition}[Token Influence Score] For a context token $j$, its influence $I_j$ is the cumulative signal magnitude weighted by its global attention reception:
\begin{equation}
    I_j = \sum_{\ell \in \mathcal{L}} \sum_{h \in \mathcal{H}} 
    \|\mathbf{v}_{j,h}^{\ell}\|_2
    \cdot 
    \sum_{i \in \mathcal{G}} A_{ij,h}^{\ell}.
    \label{eq:info_quantity}
\end{equation}
\end{definition}

\subsection{Token Pruning \& Context Realignment}
\label{sec:pruning}

\textbf{Fidelity-Constrained Pruning.}
\label{sec:adaptive_pruning}
Given the importance metric $I_j$, a fixed pruning ratio is suboptimal because the information density of the reference context varies across tasks. 
We therefore formulate token pruning as a cardinality minimization problem subject to a fidelity constraint.
It seeks the minimal subset of context tokens $\mathcal{C}' \subset \mathcal{C}$ that preserves at least a fraction $\tau$ of the total information mass $I_{\text{total}} = \sum_{j \in \mathcal{C}} I_j$:
\begin{equation}
    \min_{\mathcal{C}' \subset \mathcal{C}} |\mathcal{C}'| \quad \text{s.t.} \quad \sum_{j \in \mathcal{C}'} I_j \ge \tau \cdot I_{\text{total}},
    \label{eq:optimization}
\end{equation}
where the hyperparameter $\tau \in [0, 1]$ controls the fidelity-efficiency trade-off. 
This adaptive formulation automatically modulates the computational budget based on the semantic complexity of the task. Since the computational cost of each token is uniform, the optimal solution to Eq.~\eqref{eq:optimization} is obtained via greedy accumulation. Specifically, we sort the context tokens by their influence scores such that $I_{\pi(1)} \ge I_{\pi(2)} \ge \dots \ge I_{\pi(N)}$ and select the Top-$K$ tokens, where $K$ is the smallest integer satisfying $\sum_{m=1}^{K} I_{\pi(m)} \ge \tau \cdot I_{\text{total}}$.

\textbf{Temporal Adaptation.}
As noted in Sec.~\ref{sec:motivation}, the semantic focus of diffusion models evolves along the denoising trajectory. 
To balance adaptation precision and inference efficiency, \toolName~ adopts a piecewise-constant approximation for context selection.
Instead of recomputing token importance at every step, we define a sparse set of anchor timesteps $\mathcal{T}_{\text{anchor}} \subset \{T, \dots, 0\}$ (e.g., evenly spaced with a fixed stride). 
At anchor points, we utilize the full context to update the active context set $\mathcal{C}'_t$ for subsequent steps.
Formally, the context for the timestep $t$ is determined by:
\begin{equation}
\mathcal{C}'_t =
    \begin{cases}
        \text{Solve Eq.~\eqref{eq:optimization}}, & \text{if } t \in \mathcal{T}_{\text{anchor}} \\
        \mathcal{C}'_{t+1}, & \text{otherwise}
    \end{cases}
    \label{eq:periodic_update}
\end{equation}
This design preserves the overall trend of semantic evolution in the context (e.g., from coarse structure to fine textures), while amortizing the cost of scoring and selection across multiple steps.

\textbf{Context Realignment.}
Based on the informative subset $\mathcal{C}'_t$, we employ a hard selection strategy to compress the reference context. Note that the initial $K$ layers are exempted to preserve essential low-level feature extraction. Furthermore, addressing the positional sensitivity of Rotary Positional Embeddings (RoPE) necessitates a synchronized gather operation on both the hidden states $\mathbf{H}$ and their corresponding position indices $\mathbf{P}$:
\begin{equation}\hat{\mathbf{H}} = \mathbf{H}[\mathcal{C}'_t], \quad \hat{\mathbf{P}} = \mathbf{P}[\mathcal{C}'_t].\end{equation}
This aligned indexing mechanism ensures that the original spatial semantics and relative geometric relationships among retained tokens are preserved for subsequent layers.

\begin{figure*}[t]
    \centering
    \includegraphics[width=0.99\linewidth]{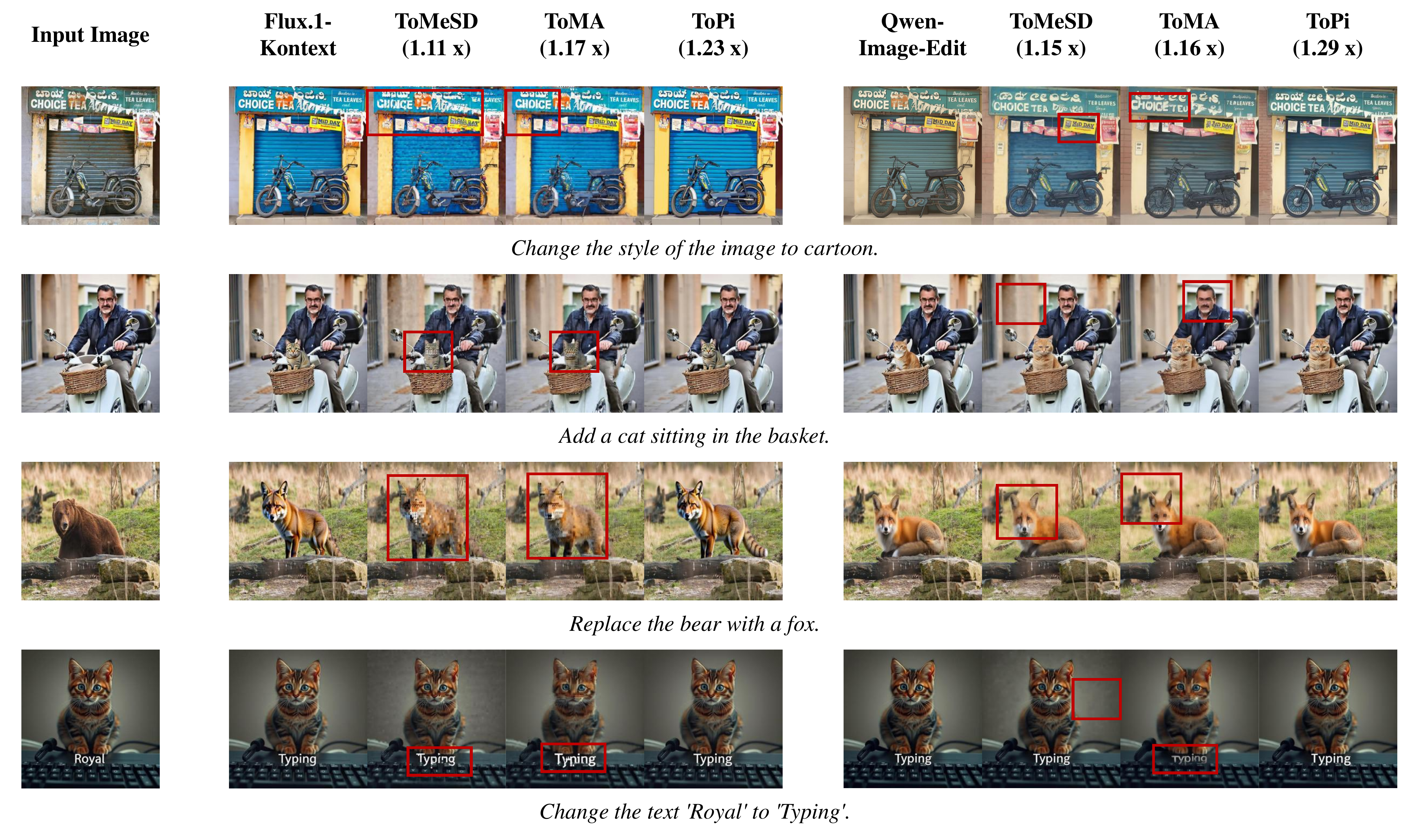}
    \caption{Qualitative comparison on the AnyEdit benchmark. We compare Flux.1-Kontext (left) against Qwen-Image-Edit (right). Best viewed zoomed in. Additional quantitative analyses are provided in Appendix~\ref{app:qualitative}. 
    }
    \label{fig:qualitative_results}
\end{figure*}

\subsection{Computational Overhead Analysis}

The overhead introduced by our framework decomposes into two components with different temporal frequencies:

\begin{itemize}
    [itemsep=6pt, topsep=0pt, parsep=0pt]
    \item \textbf{Periodic Mask Update:} : This phase includes Importance Scoring and Subset Selection, with complexities of $O(|\mathcal{L}| \cdot |\mathcal{G}| \cdot |\mathcal{C}|)$ and $O(|\mathcal{C}| \log |\mathcal{C}|)$, respectively. 
    Executed once per update interval, these costs are amortized over $\Delta T$ steps, ensuring the effective per-step overhead remains negligible.
    \item \textbf{Per-Step Pruning and Realignment:} At each diffusion timestep, we apply a global pruning operation. This reduces the instantaneous overhead to a single linear gather operation with a complexity of $O(|\mathcal{C}'|)$.
\end{itemize}

A comprehensive derivation of FLOPs and runtime benchmarks is provided in Appendix~\ref{app:flops}.

\section{Experiments}

\subsection{Experimental Setup}

\paragraph{Models and Tasks.}
We integrate {\toolName} into Flux.1-Kontext~\cite{labs2025flux1kontextflowmatching} and Qwen-Image-Edit~\cite{wu2025qwenimagetechnicalreport} to demonstrate versatility. Evaluations are conducted on the AnyEdit~\cite{yu2025anyedit} benchmark across four subtasks: (a) Camera Move Editing, (b) Global Editing, (c) Implicit Editing and (d) Local Editing.

\paragraph{Baselines.}
We benchmark {\toolName} against ToMeSD~\cite{bolya2023tomesd} and ToMA~\cite{lu2025toma}. 
To provide a comprehensive evaluation, we report results for two configurations: 
(1) \textbf{Standard}, where pruning operates globally on the entire sequence ($\mathcal{G} \cup \mathcal{C}$); and 
(2) \textbf{Reference-Only} (denoted with \textbf{*}), where we explicitly mask the noisy tokens $\mathcal{G}$ to force baselines to prune only reference tokens $\mathcal{C}$.

\paragraph{Evaluation Metrics.}
To quantitatively assess generation quality, we report three standard metrics: Peak Signal-to-Noise Ratio (PSNR) for pixel-level fidelity, Structural Similarity Index (SSIM)~\cite{wang2004image} for structural consistency (including luminance and contrast), and Learned Perceptual Image Patch Similarity (LPIPS)~\cite{zhang2018unreasonable} for perceptual similarity aligned with human vision. 
In our experiments, images generated via full-context inference serve as the ground truth.
Higher PSNR and SSIM, and lower LPIPS, indicate superior performance.

\paragraph{Implementation Details.}
Unless otherwise specified, we set the pruning threshold to $\tau = 0.85$. To balance temporal adaptability with computational overhead, the pruning mask is updated at an interval of $\Delta T = 10$.
Furthermore, we exempt the initial 10 layers from this pruning process.

\begin{table*}[t]
\centering
\caption{\textbf{Quantitative comparison across different categories.} We highlight the best results in \textbf{bold} and mark our method with a gray background. Speedup values denote the acceleration relative to the full-context inference latency. Experiments are conducted on NVIDIA A100 GPUs for Flux.1-Kontext and NVIDIA H20 GPUs for Qwen-Image-Edit, using fixed random seeds to ensure reproducibility.}
\label{tab:main_results}
\small 
\setlength{\tabcolsep}{8pt} 
\renewcommand{\arraystretch}{1.05} 
\definecolor{graybg}{gray}{0.93} 

\begin{tabular}{llcccccccc}
\toprule
 & & \multicolumn{4}{c}{\textbf{Flux.1-Kontext}} & \multicolumn{4}{c}{\textbf{Qwen-Image-Edit}} \\
\cmidrule(lr){3-6} \cmidrule(lr){7-10}
\textbf{Category} & \textbf{Method} & PSNR$\uparrow$ & SSIM$\uparrow$ & LPIPS$\downarrow$ & Speed$\uparrow$ & PSNR$\uparrow$ & SSIM$\uparrow$ & LPIPS$\downarrow$ & Speed$\uparrow$ \\
\midrule

\multirow{5}{*}{Camera Move} 
 & ToMeSD  & 20.87 & 0.678 & 0.231 & 1.05 & 23.18 & 0.737 & 0.191 & 1.14 \\
 & ToMeSD* & 21.24 & 0.685 & 0.220 & 1.10 & 22.58 & 0.724 & 0.186 & 1.17 \\
 & ToMA    & 21.79 & 0.702 & 0.213 & 1.17 & 21.33 & 0.717 & 0.221 & 1.16 \\
 & ToMA* & 21.99 & 0.695 & 0.231 & 1.18 & 21.32 & 0.717 & 0.214 & 1.16 \\
 \rowcolor{graybg} \cellcolor{white} 
 & \textbf{Ours} & \textbf{25.01} & \textbf{0.762} & \textbf{0.156} & \textbf{1.24} & \textbf{25.07} & \textbf{0.812} & \textbf{0.117} & \textbf{1.31} \\
\midrule

\multirow{5}{*}{Global Edit} 
 & ToMeSD  & 18.24 & 0.589 & 0.327 & 1.13 & 18.72 & 0.613 & 0.283 & 1.14 \\
 & ToMeSD* & 17.66 & 0.574 & 0.349 & 1.18 & 19.03 & 0.621 & 0.256 & 1.17 \\
 & ToMA    & 19.79 & 0.625 & 0.307 & 1.17 & 18.95 & 0.623 & 0.280 & 1.16 \\
 & ToMA* & 18.68 & 0.591 & 0.333 & 1.18 & 18.83 & 0.622 & 0.274 & 1.16 \\
 \rowcolor{graybg} \cellcolor{white}
 & \textbf{Ours} & \textbf{22.46} & \textbf{0.719} & \textbf{0.207} & \textbf{1.26} & \textbf{21.04} & \textbf{0.690} & \textbf{0.189} & \textbf{1.33} \\
\midrule

\multirow{5}{*}{Implicit Change} 
 & ToMeSD  & 23.33 & 0.788 & 0.178 & 1.13 & 25.56 & 0.801 & 0.135 & 1.16 \\
 & ToMeSD* & 23.06 & 0.772 & 0.189 & 1.16 & 25.66 & 0.799 & 0.124 & 1.19 \\
 & ToMA    & 24.37 & 0.793 & 0.179 & 1.18 & 21.75 & 0.774 & 0.176 & 1.17 \\
 & ToMA* & 24.39 & 0.787 & 0.187 & 1.18 & 21.71 & 0.774 & 0.169 & 1.17 \\
 \rowcolor{graybg} \cellcolor{white}
 & \textbf{Ours} & \textbf{29.40} & \textbf{0.881} & \textbf{0.085} & \textbf{1.21} & \textbf{28.05} & \textbf{0.848} & \textbf{0.085} & \textbf{1.27} \\
\midrule

\multirow{5}{*}{Local Edit} 
 & ToMeSD  & 21.36 & 0.743 & 0.203 & 1.13 & 23.33 & 0.763 & 0.158 & 1.15 \\
 & ToMeSD* & 21.65 & 0.748 & 0.193 & 1.17 & 22.90 & 0.753 & 0.153 & 1.18 \\
 & ToMA    & 22.69 & 0.772 & 0.178 & 1.17 & 21.48 & 0.738 & 0.185 & 1.16 \\
 & ToMA* & 22.43 & 0.756 & 0.195 & 1.18 & 21.46 & 0.738 & 0.177 & 1.16 \\
 \rowcolor{graybg} \cellcolor{white}
 & \textbf{Ours} & \textbf{27.55} & \textbf{0.868} & \textbf{0.087} & \textbf{1.21} & \textbf{25.82} & \textbf{0.823} & \textbf{0.103} & \textbf{1.27} \\
\bottomrule
\end{tabular}
\end{table*}

\subsection{Main Results}
To evaluate the efficacy of \toolName, we conduct a comprehensive benchmarking analysis encompassing both qualitative visual fidelity and quantitative performance metrics.

\subsubsection{Qualitative results}
Existing token reduction methods (e.g., ToMeSD and ToMA) often compromise generation fidelity, manifesting as perceptual blurring artifacts or semantic drift from the reference. As visually corroborated in Fig.~\ref{fig:qualitative_results}, these baseline methods struggle to retain fine-grained textures under high compression regimes, leading to structural deterioration. Representative instances of this degradation include the illegible text in the garage scene (Row 1) and the over-smoothed fur texture in the animal replacement task (Row 3). Conversely, \toolName{} demonstrates superior perceptual fidelity. It effectively preserves critical high-frequency details and produces outputs that remain visually close to the computationally intensive full-context baseline. 

\subsubsection{Quantitative Results}
Tab.~\ref{tab:main_results} reports the quantitative evaluation on Flux.1-Kontext and Qwen-Image-Edit models. Our framework consistently achieves state-of-the-art performance, outperforming baselines by substantial margins across all metrics and tasks. Specifically, in tasks demanding high structural fidelity such as \textit{Implicit Change}, our method yields a remarkable gain of over 5 in PSNR compared to ToMA (29.40 vs. 24.37 on Flux.1-Kontext). Similarly, in the \textit{Local Editing} scenario, we minimize the LPIPS to 0.087, effectively halving the error relative to the best-performing baseline (0.178). 

Importantly, these quality gains do not come at the cost of latency. As detailed in Tab.~\ref{tab:main_results}, our approach consistently establishes a superior quality–latency Pareto frontier. Unlike baselines that compromise visual quality for moderate acceleration (1.05$\times$--1.18$\times$), our method achieves the highest inference throughput, reaching speedups of up to 1.33$\times$ on Qwen-Image-Edit. 
This efficiency stems from our lightweight pruning paradigm, which effectively eliminates redundancy in reference tokens with negligible run-time computational overhead.

Notably, we also compare against ``Reference-Only'' baseline variants (marked with \textbf{*}), where pruning is manually constrained to reference tokens $\mathcal{C}$ to favor preservation. 
Even with this adaptation, our method consistently outperforms them (e.g., 25.01 vs. 21.24 in \textit{Camera Move}), proving that our dedicated selection mechanism is far more effective than simply masking existing pruning algorithms.

\subsection{Orthogonality with Temporal Acceleration}
\begin{figure}[t]
    \centering
    \includegraphics[width=1.0\linewidth]{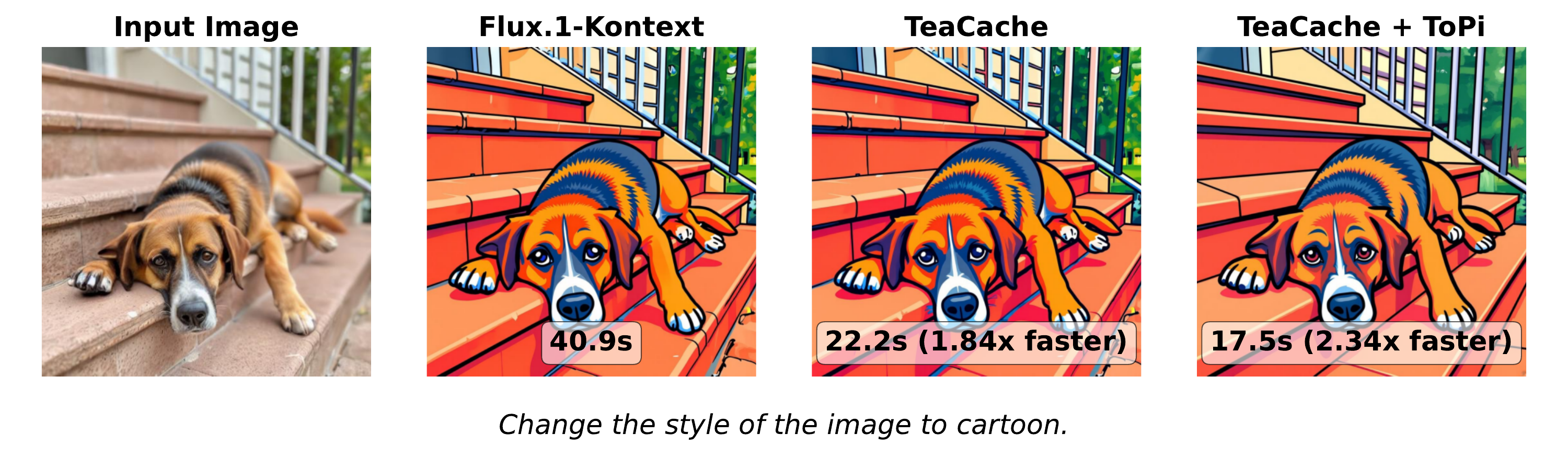}
    \caption{Qualitative demonstration of the orthogonality between {\toolName} and TeaCache. {\toolName} preserves high-precision semantic details even when combined with TeaCache.}
    \label{fig:tea}
\end{figure}

To verify the orthogonality with temporal acceleration techniques, we deployed {\toolName} alongside TeaCache~\cite{liu2025timestep}. Empirical results show that the combination achieves a cumulative speedup of 2.34$\times$. 
Qualitatively, as shown in Fig.~\ref{fig:tea}, {\toolName} complements the temporal acceleration of TeaCache by acting as a high-precision spatial filter that preserves semantic details even when features are reused across timesteps, offering a superior trade-off compared to using either method alone.

\subsection{Ablation Study}
\label{sec:ablation}

In this section, we conduct comprehensive ablation studies to validate the effectiveness of key components in {\toolName}. To ensure a consistent and rigorous evaluation, we utilize the Flux.1-Kontext model as our primary testbed and standardize the token pruning ratio at 50\% (implemented via a static Top-K selection) across all experiments. This strict constraint serves as a stress test, allowing us to isolate the contribution of each design choice, specifically the layer selection strategy and the importance metric definition, independent of the adaptive mechanism.
Additionally, Appendix~\ref{app:ablation} provides further analyses, including the impact of temporal adaptation, a comparison between token reduction strategies, and the effect of the pruning threshold $\tau$.

\begin{table}[h]
\centering
\caption{Ablation study on \textbf{Layer Selection Strategies}.}
\renewcommand{\arraystretch}{1.05} 
\definecolor{graybg}{gray}{0.93} 
\label{tab:ablation_layer_selection}
\small
\setlength{\tabcolsep}{5pt} 
\begin{tabular}{ll ccc}
\toprule
Category & Method & PSNR $\uparrow$ & SSIM $\uparrow$ & LPIPS $\downarrow$ \\
\midrule
\multirow{3}{*}{Camera Move} 
    & Random  & 26.80 & 0.819 & 0.112 \\
    & All & 27.43 & 0.822 & 0.113 \\
    \rowcolor{graybg} \cellcolor{white}
    & \textbf{Ours} & \textbf{27.69} & \textbf{0.828} & \textbf{0.108} \\
\midrule
\multirow{3}{*}{Global Editing} 
    & Random  & 23.13 & 0.740 & 0.184 \\
    & All & 23.64 & 0.751 & 0.176 \\
    \rowcolor{graybg} \cellcolor{white}
    & \textbf{Ours} & \textbf{23.76} & \textbf{0.754} & \textbf{0.173} \\
\midrule
\multirow{3}{*}{Implicit Change} 
    & Random  & 29.31 & 0.868 & 0.090 \\
    & All & 30.00 & 0.870 & 0.091 \\
    \rowcolor{graybg} \cellcolor{white}
    & \textbf{Ours} & \textbf{30.02} & \textbf{0.874} & \textbf{0.088} \\
\midrule
\multirow{3}{*}{Local Editing} 
    & Random  & 26.62 & 0.849 & 0.099 \\
    & All & 27.48 & 0.854 & 0.097 \\
    \rowcolor{graybg} \cellcolor{white}
    & \textbf{Ours} & \textbf{27.52} & \textbf{0.862} & \textbf{0.091} \\
\bottomrule
\end{tabular}
\end{table}

\paragraph{Layer Selection Strategy.}
To evaluate the layer selection method proposed in Sec.~\ref{sec:layer_selection}, we compare our context sensitivity score $S^{(\ell)}$ with two baselines: (1) \textit{Random}, which uniformly samples $3$ layers; and (2) \textit{All}, which aggregates scores from all layers. As shown in Tab.~\ref{tab:ablation_layer_selection}, our strategy performs matches or slightly exceeds the \textit{All} baseline. This indicates that using every layer introduces noise from irrelevant layers, which weakens the important semantic signals. By selecting the layers with the highest $S^{(\ell)}$, our method ensures that token importance is derived only from the most useful interactions, allowing us to achieve robust results using a smaller subset of layers.

\begin{table}[htbp]
\centering
\caption{Ablation study on \textbf{Token Pruning Metrics}.}
\renewcommand{\arraystretch}{1.05} 
\definecolor{graybg}{gray}{0.93} 
\label{tab:ablation_pruning_metric}
\small
\setlength{\tabcolsep}{5pt}
\begin{tabular}{ll ccc}
\toprule
Category & Method & PSNR $\uparrow$ & SSIM $\uparrow$ & LPIPS $\downarrow$ \\
\midrule
\multirow{3}{*}{Camera Move} 
    & Similarity    & 17.21 & 0.657 & 0.292 \\
    & Attn-Only     & 27.39 & 0.825 & 0.111 \\
    \rowcolor{graybg} \cellcolor{white}
    & \textbf{Ours} & \textbf{27.69} & \textbf{0.828} & \textbf{0.108} \\
\midrule
\multirow{3}{*}{Global Editing} 
    & Similarity    & 17.06 & 0.629 & 0.312 \\
    & Attn-Only     & 23.60 & 0.751 & \textbf{0.173} \\
    \rowcolor{graybg} \cellcolor{white}
    & \textbf{Ours} & \textbf{23.76} & \textbf{0.754} & \textbf{0.173} \\
\midrule
\multirow{3}{*}{Implicit Change} 
    & Similarity    & 17.64 & 0.754 & 0.265 \\
    & Attn-Only     & 29.41 & 0.867 & 0.092 \\
    \rowcolor{graybg} \cellcolor{white}
    & \textbf{Ours} & \textbf{30.02} & \textbf{0.874} & \textbf{0.088} \\
\midrule
\multirow{3}{*}{Local Editing} 
    & Similarity    & 17.32 & 0.720 & 0.251 \\
    & Attn-Only     & 27.03 & 0.856 & 0.095 \\
    \rowcolor{graybg} \cellcolor{white}
    & \textbf{Ours} & \textbf{27.52} & \textbf{0.862} & \textbf{0.091} \\
\bottomrule
\end{tabular}
\end{table}

\paragraph{Pruning Metric.}
\label{sec:ablation_metric}
We evaluate the \textit{Token Influence Score} (Eq.~\ref{eq:info_quantity}) against two baselines: (1) \textit{Feature Similarity}, ranking tokens by cosine similarity; and (2) \textit{Attn-Only}, relying solely on the attention map $\mathbf{A}$. Tab.~\ref{tab:ablation_pruning_metric} shows that \textit{Feature Similarity} degrades performance, indicating semantic alignment is a poor proxy for functional importance. While \textit{Attn-Only} is competitive, our method achieves superior fidelity by incorporating the value magnitude $\|\mathbf{V}\|$. This confirms that attention weights alone are insufficient; influential tokens must possess both high attention and significant magnitude. By integrating $\|\mathbf{V}\|$, we remove spurious high-attention tokens to capture the most critical feature updates.

\section{Conclusion}

In this paper, we introduce {\toolName}, a novel training-free framework designed to accelerate in-context generation within Diffusion Transformers. Through a systematic analysis of spatiotemporal attention dynamics, we identify that the processing of contextual reference tokens is characterized by significant redundancy. Departing from prior heuristic-based pruning, {\toolName} employs an influence-aware metric derived directly from singal contribution preservation. This formulation enables the precise identification and pruning of low-utility tokens while maintaining the semantic fidelity of the generation process. Extensive evaluations across diverse architectures demonstrate that {\toolName} establishes a superior trade-off between inference latency and generation quality, delivering speedups with negligible perceptual degradation. 
By demonstrating that sparse token retention is sufficient for robust in-context generation, this work facilitates efficient model deployment and highlights a promising direction for future input-adaptive generative models.
\newpage
\section*{Impact Statement}

This paper presents work whose goal is to advance the field of Machine
Learning. There are many potential societal consequences of our work, none
which we feel must be specifically highlighted here.

\bibliography{ref}
\bibliographystyle{icml2026}

\newpage
\appendix
\onecolumn

\begin{figure*}
    \centering
    \includegraphics[width=0.99\linewidth]{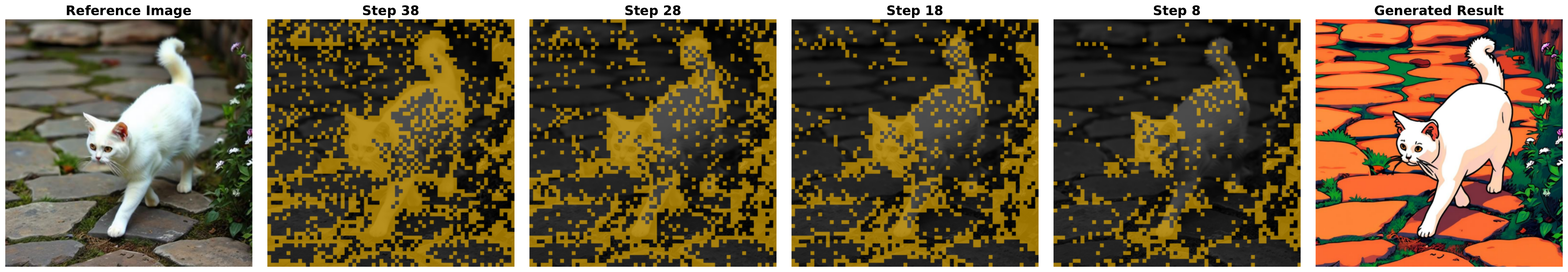}
    \caption{\textbf{Visualization of token selection dynamics across the diffusion trajectory.}
    We examine pruning outcomes at four distinct denoising stages: Early ($t=38$), Middle ($t=28, 18$), and Late ($t=8$). The gold binary mask overlaid on the reference image represents the top-$\tau$ ($\tau=0.85$) cumulative probability mass, highlighting the retained semantic anchors versus the pruned background.}
    \label{fig:hotmap}
\end{figure*}

\section{Visualization of Token Selection Dynamics}
\label{sec:appendix_temporal_vis}
This section empirically validates the precision of the \toolName~mechanism and underscores the necessity of our periodic update strategy. Fig.~\ref{fig:hotmap} visualizes the evolution of the active token set throughout the denoising process.
The visualization reveals a progressive sparsification of the active token set. As the denoising trajectory resolves latent structure, the model's reliance shifts from external reference guidance to internal self-refinement. In the early stages, the generation necessitates broad semantic coverage from the reference to establish global structure. However, as the image takes shape, the generative process becomes increasingly driven by self-attention within the noisy tokens, reducing the need for extensive contextual support. Consequently, the cross-attention mechanism naturally discards the majority of reference tokens, retaining only sparse semantic anchors.

\begin{algorithm}[ht]
   \caption{Inference with \toolName}
   \label{alg:main_method}
\begin{algorithmic}[1]
   \STATE {\bfseries Input:}  Prompt $\text{y}$, Noisy tokens $\mathcal{G}^t$, Reference tokens $\mathcal{C}$, RoPE $\mathcal{P}$, \\
   Update Timesteps $\mathcal{T_{\text{anchor}}}$, Representative Layers $\mathcal{L}$, Ratio $\tau$
   \STATE {\bfseries Output:} Generated Image $I_{gen}$
   
   \STATE \textcolor{gray}{// Initialize: Start with full token retention}
   \STATE $M \leftarrow \mathbf{1}$ 
   
   \FOR{$t = T$ {\bfseries to} $1$}
       \STATE \textcolor{gray}{// 1. Common Front-End (Full Context)}
       \STATE $\mathbf{H}_G, \mathbf{H}_C, \mathbf{H}_y \leftarrow \text{DiT}_{\text{front}}(\mathcal{G}^t, \mathcal{C}, \text{y}, \mathcal{P})$

       \IF{$t \in \mathcal{T_{\text{anchor}}}$}
           \STATE \textcolor{gray}{// --- Mode A: Full Inference \& Mask Update ---}
           \STATE \textcolor{gray}{// Run with full reference to capture attention maps}
           \STATE $\epsilon_\theta, \mathbf{A}_{\mathcal{L}}, \mathbf{V}_{\mathcal{L}} \leftarrow \text{DiT}_{\text{back}}(\mathbf{H}_G, \mathbf{H}_C, \mathbf{H}_y,\mathcal{P}, \text{return\_attn}=\text{True}, \text{anchor\_layers}=\mathcal{L})$
           
           \STATE \textcolor{gray}{// Update mask for future steps}
           \STATE $\mathcal{I} \leftarrow \text{Aggregate}(\mathbf{A}_{\mathcal{L}},  \mathbf{V}_{\mathcal{L}})$
        \STATE $\mathcal{I}_{\pi}, \text{Indices} \leftarrow \text{SortDescending}(\mathcal{I})$ 
        
        \STATE $K \leftarrow \arg\min_k (\sum_{j=1}^k \mathcal{I}_{\pi}[j] \ge \tau \cdot I_{total})$
        \STATE $M \leftarrow \text{CreateMask}(\text{Indices}[0:K])$
       \ELSE
           \STATE \textcolor{gray}{// --- Mode B: Accelerated Inference ---}
           \STATE \textcolor{gray}{// Physically remove uninformative tokens (Hard Pruning)}
           \STATE $\hat{\mathbf{H}}_C \leftarrow \mathbf{H}_C[M]$ 
           \STATE $\hat{\mathcal{P}} \leftarrow cat(\mathcal{P}[\mathcal{G}], \mathcal{P}[M])$ 
           
           \STATE \textcolor{gray}{// Run backend with reduced sequence length}
           \STATE $\epsilon_\theta \leftarrow \text{DiT}_{\text{back}}(\mathbf{H}_G, \hat{\mathbf{H}}_C, \mathbf{H}_y,\mathcal{\hat{P}} , \text{return\_attn}=\text{False})$
       \ENDIF

       \STATE \textcolor{gray}{// 2. Standard Diffusion Step}
       \STATE $\mathcal{G}_{t-1} \leftarrow \text{Step}(\mathcal{G}^t, \epsilon_\theta)$
   \ENDFOR
   
   \STATE $I_{gen} \leftarrow \text{Decode}(\mathcal{G}_0)$
\end{algorithmic}
\end{algorithm}

\section{Detailed Algorithmic Formulation of \toolName}
\label{app:algo}

Algorithm~\ref{alg:main_method} delineates the inference procedure of \toolName. To facilitate token-wise optimization, we conceptually partition the diffusion backbone (DiT) into a frontend ($\text{DiT}_{\text{front}}$) and a backend ($\text{DiT}_{\text{back}}$). Specifically, $\text{DiT}_{\text{front}}$ comprises the first $K$ layers, while $\text{DiT}_{\text{back}}$ encompasses the remaining layers. The inference trajectory alternates between two distinct operational modes, governed by the update timesteps $\mathcal{T}_\text{anchor}$:

\begin{itemize}
    \item \textbf{Mask Update Step ($t \in \mathcal{T}_\text{anchor}$):} During these denoising steps, the model executes a full-reference inference pass via $\text{DiT}_{\text{back}}$ to extract precise inter-token cross-attention maps $\mathbf{A}$. These maps are aggregated to quantify token saliency, which subsequently recalibrates the binary selection mask $M$ for future iterations.
    \item \textbf{Accelerated Step ($t \notin \mathcal{T}_\text{anchor}$):} During these denoising steps, the pre-computed mask $M$ is applied to physically prune redundant reference tokens (Hard Pruning). This yields a compressed latent representation $\hat{H}_C$, allowing $\text{DiT}_{\text{back}}$ to operate on a significantly reduced sequence length, thereby minimizing computational overhead without compromising semantic fidelity.
\end{itemize}

\section{Implementation Details}
\label{app:details}
\subsection{Model Architecture}
\begin{itemize}

\item \textbf{Flux.1-Kontext}~\cite{labs2025flux1kontextflowmatching}. Flux.1-Kontext  is a 12B parameter rectified flow transformer specialized for text-instruction-based image editing. Based on the FluxTransformer2DModel architecture, it incorporates explicit guidance embeddings and features a hybrid layer structure consisting of 19 dual-stream attention layers and 38 single-stream layers. 

\item \textbf{Qwen-Image-Edit}~\cite{wu2025qwenimagetechnicalreport}. Qwen-Image-Edit is an image editing model built upon the 20B parameter Qwen-Image foundation, designed to integrate text rendering capabilities with precise semantic and appearance control via Qwen2.5-VL and a VAE encoder. Architecturally, it utilizes the QwenImageTransformer2DModel structure, featuring a deep configuration of 60 transformer layers.

\end{itemize}

\subsection{Hardware Environment} 

All evaluations were conducted on:
\begin{itemize}
    \item \textbf{Flux.1-Kontext:} NVIDIA A100 (40GB) GPU.
    \item \textbf{Qwen-Image-Edit:} NVIDIA H20 (96GB) GPU.
    \item \textbf{Software:} PyTorch 2.4 with FlashAttention enabled for memory efficient attention. 
\end{itemize}

\subsection{Calibration}
To ensure the robustness and generality of the identified Representative Layers ($\mathcal{L}$), we constructed a diverse calibration set of 48 image-text pairs. Crucially, these samples were stratified across varying subtasks (e.g., Local Editing, Global Editing) to avoid task-specific bias. We observed that the distribution of layer sensitivity remained consistent across these subtasks, supporting our hypothesis that layer importance is structurally intrinsic.
Meanwhile, calibration is a one-time offline procedure per architecture, incurring zero runtime overhead during inference. Specifically, we identified $\mathcal{L} = \{13, 18, 31\}$ for Flux.1-Kontext and $\mathcal{L} = \{35, 37, 42\}$ for Qwen-Image-Edit.

\begin{figure}[]
    \centering
    \includegraphics[width=0.95\linewidth]{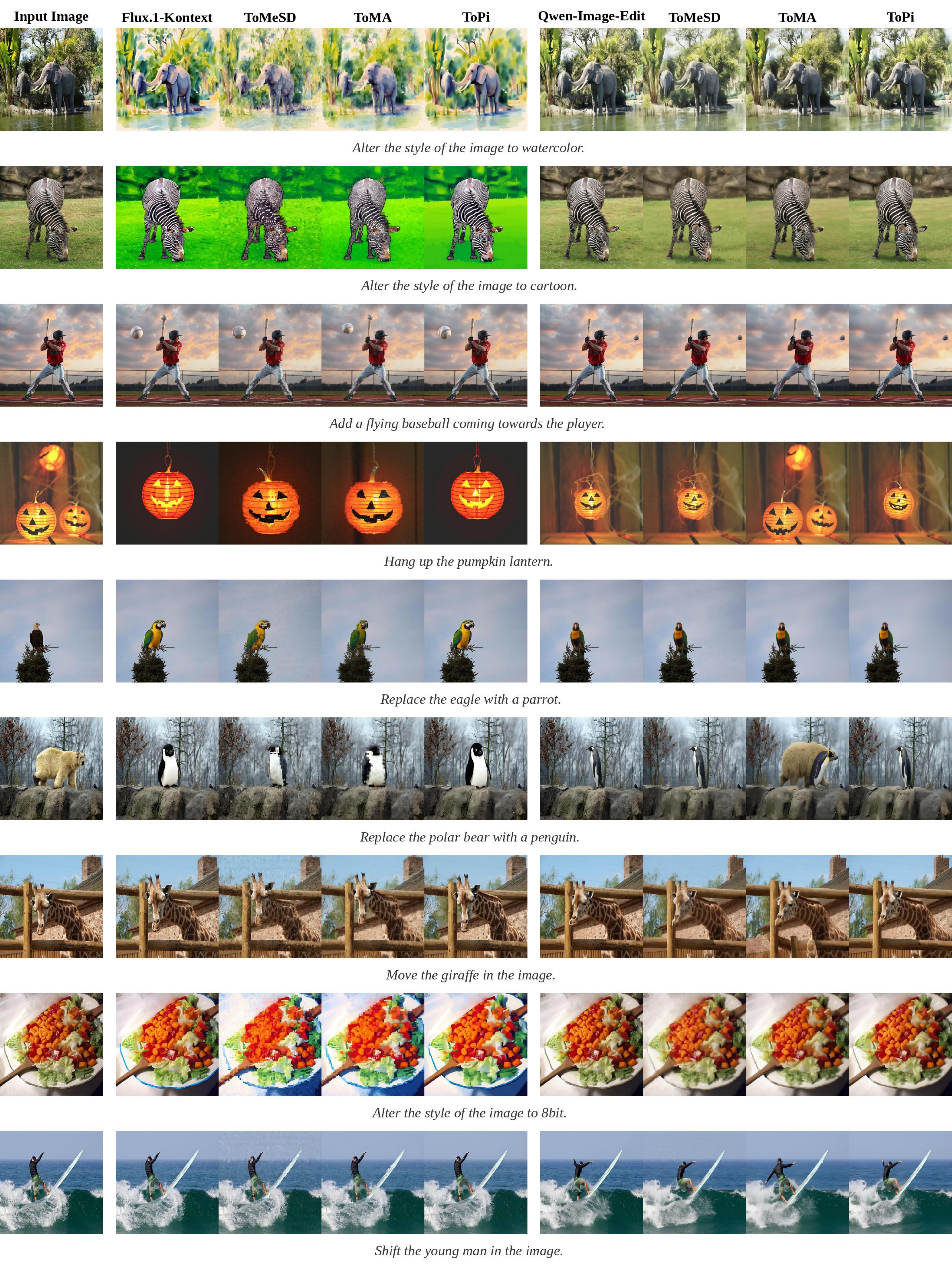}
    \caption{Qualitative comparison against ToMeSD and ToMA on the AnyEdit benchmark. The visualization is bilaterally partitioned: the left panel utilizes the \textbf{Flux.1-Kontext} backbone, while the right panel employs \textbf{Qwen-Image-Edit}. Best viewed zoomed in.}
    \label{fig:supp_visuals}
\end{figure}

\section{More Qualitative Results} \label{app:qualitative}
To further evaluate the robustness and generalization ability of \toolName{}, we provide extended qualitative comparisons across a broader range of editing scenarios. 

\subsection{Single Image}
Due to space constraints in the main text, we present additional qualitative comparisons here. 
Fig.~\ref{fig:supp_visuals} demonstrates the performance of \toolName{} across a wider variety of tasks, including complex scene composition and style transfer. As observed, our method maintains high fidelity to the reference subject, whereas baseline methods often lose fine-grained details or exhibit artifacts in texture generation.

\subsection{Multiple Image}

\begin{figure}[]
    \centering
    \includegraphics[width=0.99\linewidth]{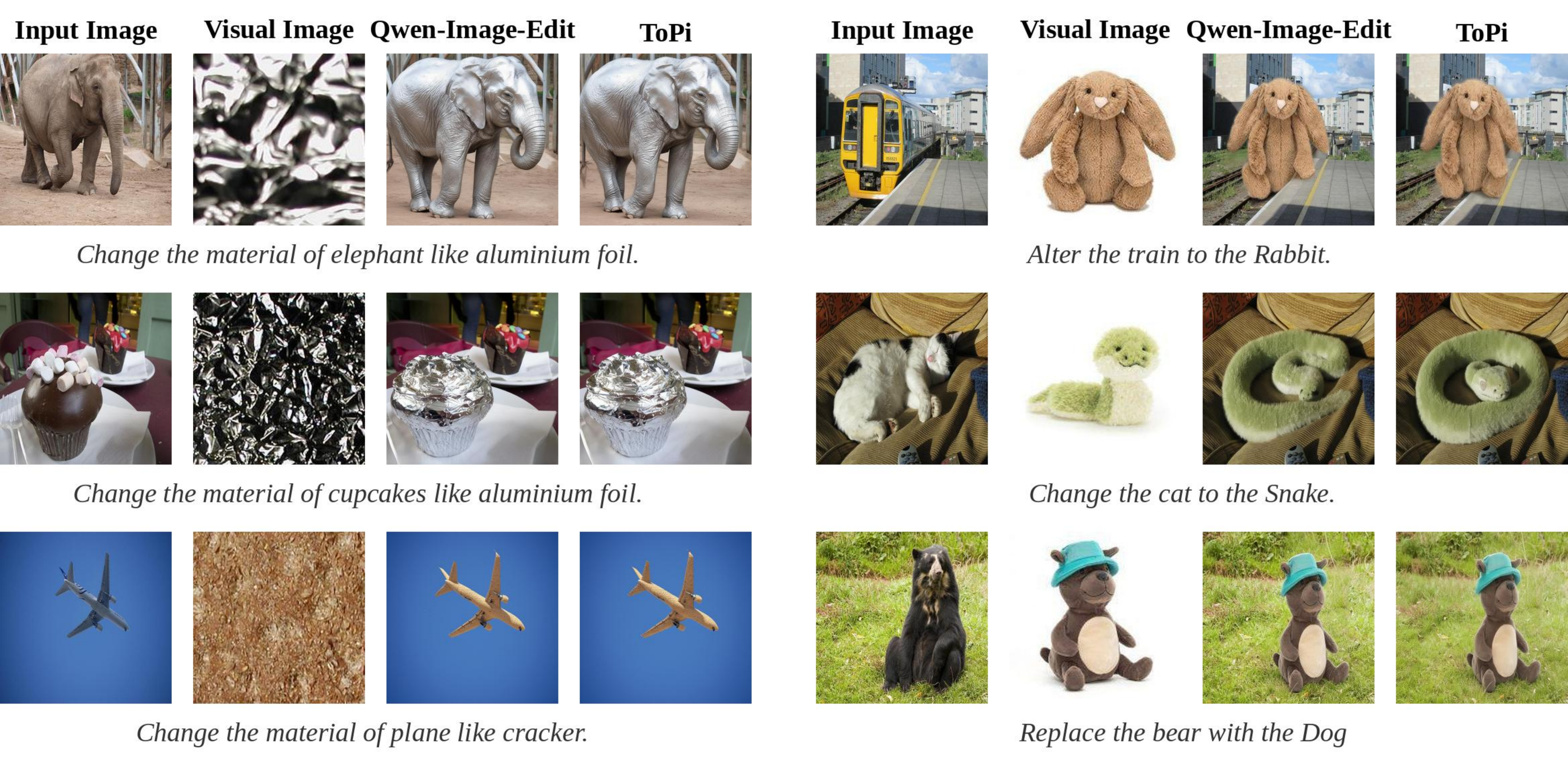}
    \caption{Qualitative comparison on multi-image reference editing tasks. }
    \label{fig:multi_task_vis}
\end{figure}

We extend our evaluation to multi-image scenarios using the Qwen-Image-Edit architecture. Fig.~\ref{fig:multi_task_vis} compares our pruned model (\toolName{}) against the full-context inference baseline across tasks like visual material transfer and reference-guided object replacement.
As observed, \toolName{} maintains high generation fidelity despite the token pruning. It successfully preserves structural integrity during texture mapping and maintains object identity during reference-guided replacement. These results confirm that our metric accurately identifies and retains the critical semantic tokens required for complex, multi-input generation, achieving performance parity with the full-context baseline.

\begin{figure}
    \centering
    \includegraphics[width=0.9\linewidth]{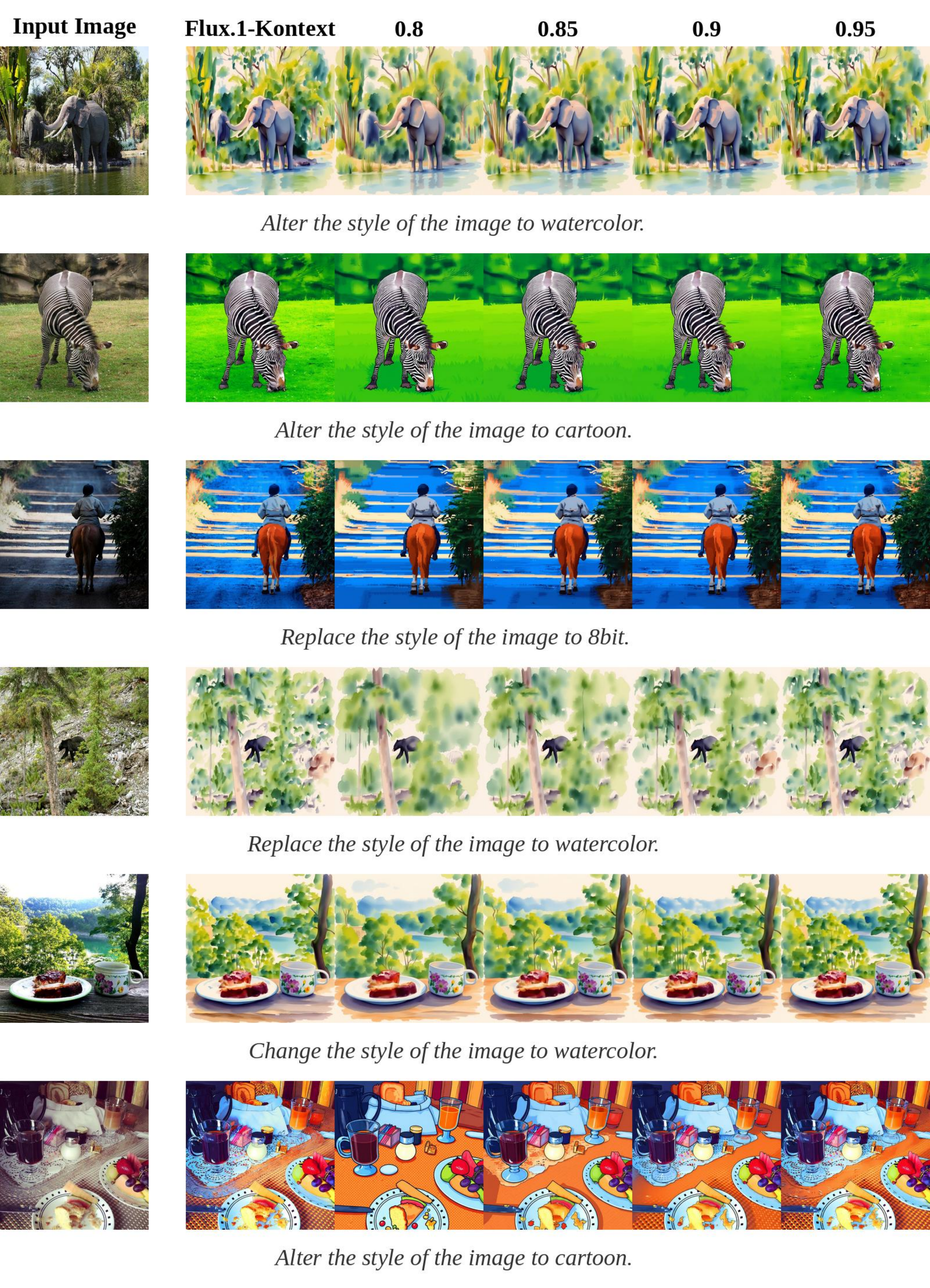}
    \caption{Qualitative visualization of the generated samples across varying $\tau$ values.}
    \label{fig:comp_ratio}
\end{figure}

\section{More Ablation Results} \label{app:ablation}

\subsection{Impact of Pruning Threshold $\tau$}
We investigate the sensitivity of \toolName{} to the pruning threshold $\tau$. Fig.~\ref{fig:comp_ratio} provides a qualitative visualization of the generated samples across varying $\tau$ values. As observed, our method maintains high visual fidelity and structural consistency even under aggressive pruning rates (e.g., $\tau=0.8$). The results demonstrate that \toolName{} effectively removes redundant tokens without compromising the semantic integrity of the generated images.

\newpage

\subsection{Prune vs Merge}

\begin{table}[h]
\centering
\caption{Quantitative comparison across different editing categories. The best results are highlighted in \textbf{bold}.}
\label{tab:merged_comparison_wide}
\setlength{\tabcolsep}{4pt} 
\begin{tabular}{lcccccc}
\toprule
 & \multicolumn{2}{c}{PSNR $\uparrow$} & \multicolumn{2}{c}{SSIM $\uparrow$} & \multicolumn{2}{c}{LPIPS $\downarrow$} \\
\cmidrule(lr){2-3} \cmidrule(lr){4-5} \cmidrule(lr){6-7}
Category & Prune & Merge & Prune & Merge & Prune & Merge \\
\midrule
Camera Move & \textbf{27.69} & 26.12 & \textbf{0.828} & 0.799 & \textbf{0.108} & 0.136 \\
Global Edit & \textbf{23.76} & 21.84 & \textbf{0.754} & 0.708 & \textbf{0.173} & 0.224 \\
Implicit Change & \textbf{30.02} & 28.27 & \textbf{0.874} & 0.850 & \textbf{0.088} & 0.119 \\
Local Edit & \textbf{27.52} & 25.75 & \textbf{0.862} & 0.830 & \textbf{0.091} & 0.121 \\
\bottomrule
\end{tabular}
\end{table}

Tab. \ref{tab:merged_comparison_wide} provides a quantitative ablation study comparing two distinct token reduction strategies within our DiT framework: \textit{Prune} (selective retention) and \textit{Merge} (token aggregation). To ensure a fair comparison, the \textit{Merge} baseline utilizes the same Token Influence Score ($I_j$) for ranking. However, instead of discarding the less important tokens, we aggregate them into their nearest retained neighbors via feature averaging.
The results indicate that the merging strategy consistently lags behind the pruning approach across all editing categories. We hypothesize that merging tokens inevitably introduces feature over-smoothing, where distinct high-frequency details are averaged out, leading to higher LPIPS values (0.224 vs. 0.173 in Global Editing). Conversely, \textit{Prune} preserves the original feature integrity of the most critical tokens, thereby achieving superior fidelity.

\subsection{Comparison with Static Layer Selection.}

\begin{table}[h]
\centering
\caption{\textbf{Quantitative comparison of layer selection strategies.} We benchmark our adaptive method against static heuristics (\textit{First}, \textit{Last}) and aggregation baselines (\textit{Random}, \textit{All}).}
\label{tab:ablation_layer_selection_app}
\small
\setlength{\tabcolsep}{5pt}
\begin{tabular}{ll ccc}
\toprule
Category & Selection Strategy & PSNR $\uparrow$ & SSIM $\uparrow$ & LPIPS $\downarrow$ \\
\midrule
\multirow{5}{*}{Camera Move} 
    & First   & 27.10 & 0.817 & 0.118 \\
    & Last    & 24.31 & 0.772 & 0.149 \\
    & Random  & 26.80 & 0.819 & 0.112 \\
    & All & 27.43 & 0.822 & 0.113 \\
    & \textbf{Ours} & \textbf{27.69} & \textbf{0.828} & \textbf{0.108} \\
\midrule
\multirow{5}{*}{Global Editing} 
    & First   & 22.76 & 0.731 & 0.195 \\
    & Last     & 22.13 & 0.726 & 0.198 \\
    & Random  & 23.13 & 0.740 & 0.184 \\
    & All & 23.64 & 0.751 & 0.176 \\
    & \textbf{Ours} & \textbf{23.76} & \textbf{0.754} & \textbf{0.173} \\
\midrule
\multirow{5}{*}{Implicit Change} 
    & First   & 29.35 & 0.866 & 0.097 \\
    & Last    & 27.55 & 0.857 & 0.107 \\
    & Random  & 29.31 & 0.868 & 0.090 \\
    & All & 30.00 & 0.870 & 0.091 \\
    & \textbf{Ours} & \textbf{30.02} & \textbf{0.874} & \textbf{0.088} \\
\midrule
\multirow{5}{*}{Local Editing} 
    & First   & 26.93 & 0.845 & 0.104 \\
    & Last    & 25.21 & 0.832 & 0.119 \\
    & Random  & 26.62 & 0.849 & 0.099 \\
    & All & 27.48 & 0.854 & 0.097 \\
    & \textbf{Ours} & \textbf{27.52} & \textbf{0.862} & \textbf{0.091} \\
\bottomrule
\end{tabular}
\end{table}

Complementing the discussion on \textit{Random} and \textit{All} strategies in the main text, here we further benchmark our method against two deterministic heuristics: (1) \textit{First}, which selects the initial $3$ layers; and (2) \textit{Last}, targeting the final $3$ layers.
As detailed in Tab.~\ref{tab:ablation_layer_selection_app}, two key observations support the necessity of our adaptive selection mechanism. 
\textbf{First, static heuristics are insufficient.} The \textit{Last} strategy yields the lowest performance across all categories (e.g., PSNR 24.31 vs. 27.69 in Camera Move), suggesting that the deepest layers in DiT may focus more on high-frequency noise prediction rather than semantic context alignment. While \textit{First} performs better, it still lags behind \textit{Ours}, indicating that critical information is not strictly confined to the shallowest layers.
\textbf{Second, representative layer anchoring is crucial.} Unlike these fixed strategies, our method dynamically isolates layers with high context sensitivity scores $S^{(\ell)}$. This effectively filters out non-informative layers (avoiding the noise introduction seen in \textit{All}) while capturing essential semantic features that static strategies miss.

\begin{figure}[t]
    \centering
    \includegraphics[width=0.95\linewidth]{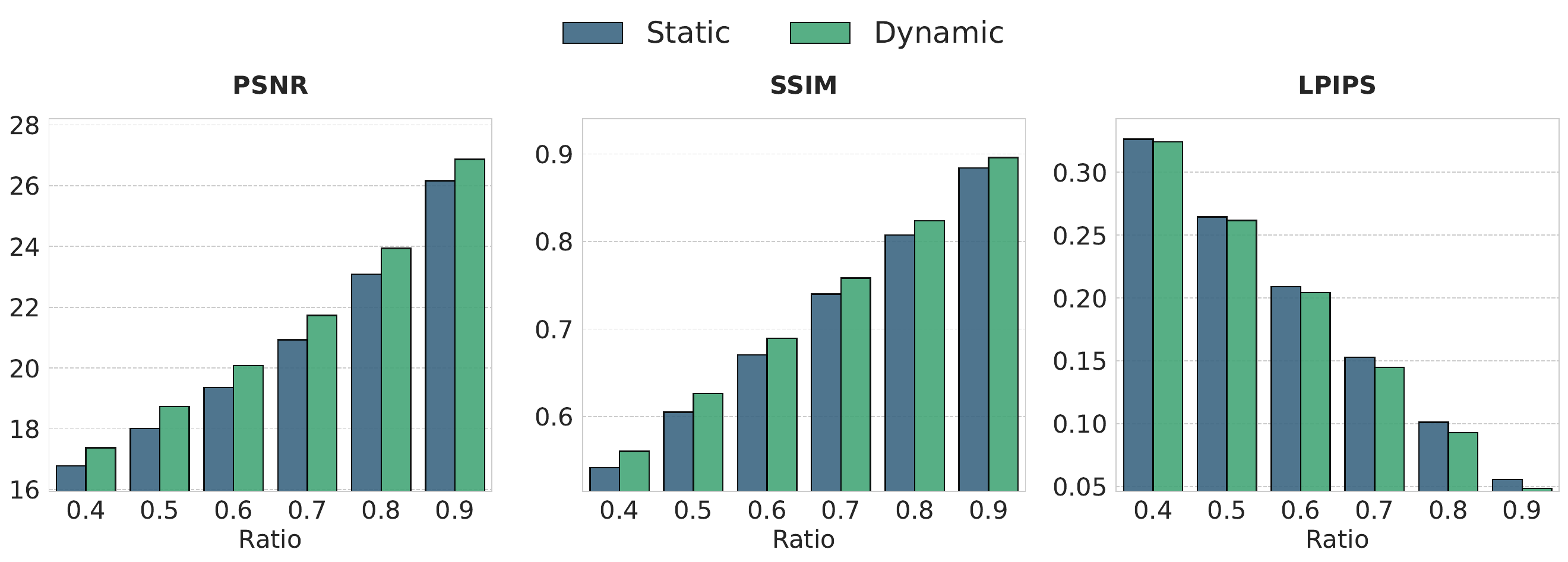}
    \caption{Effectiveness of Temporal Adaptation}
    \label{fig:dynamic_comparison}
\end{figure}

\subsection{Impact of Temporal Adaptation}
\label{sec:ablation_temporal}

We validate the necessity of the token mask update (Eq.~\ref{eq:periodic_update}) by comparing \toolName~ against a \textit{Static} baseline where the context set is determined once at initialization. As shown in Fig.~\ref{fig:dynamic_comparison}, our dynamic strategy consistently outperforms the static approach across all metrics (PSNR, SSIM, LPIPS). This confirms that a fixed token subset fails to accommodate the evolving semantic focus of the diffusion process (e.g., from structure to texture), whereas our temporal adaptation effectively captures these critical shifts.

\section{Detailed Complexity and FLOPs Analysis} \label{app:flops}

We provide a comprehensive theoretical analysis of the computational complexity and a breakdown of the FLOPs reduction achieved by \toolName{}.

\subsection{Theoretical Complexity Analysis: Joint vs. Single Blocks}

Based on the official implementation of Flux.1-Kontext and Qwen-Image-Edit, we derive the computational complexity for its two distinct block types: Double Stream Block and Single Stream Block. 
Based on the \texttt{QwenDoubleStreamAttnProcessor} implementation, Qwen-Image-Edit employs a transformer block mechanism similar to the Flux Double Stream Block, operating at a large scale (e.g., hidden dimension $d=4096$ and substantial layer number $L = 60$).
Let $N_{img}$ be the number of visual tokens, $N_{prompt}$ the number of prompt text tokens, and $d$ the hidden dimension.

\textbf{1. Double Stream Block.}
This block processes streams independently for linear projections but jointly for attention.
\begin{itemize}
    \item \textbf{Joint Attention}: 
    \begin{equation}
        \mathcal{C}_{JointAttn} \approx \underbrace{4 (N_{img} + N_{prompt})^2 d}_{\text{Attention Core}} 
    \end{equation}
    \item \textbf{Dual FFNs}: 
        \begin{itemize}
            \item Flux.1-Kontext: Separate MLPs ($d \to 4d \to d$) are applied to each stream.
                \begin{equation}
                    \mathcal{C}_{DualFFN} \approx \underbrace{16 N_{img} d^2}_{\text{Image FFN}} + \underbrace{16 N_{prompt} d^2}_{\text{Text FFN}} + \underbrace{8 (N_{img} + N_{prompt}) d^2}_{\text{Q,K,V,O Projections}} = 24 (N_{img} + N_{prompt}) d^2
                \end{equation}
            \item Qwen-Image-Edit: Separate MLPs ($d \to 6d \to d$) are applied to each stream.
                \begin{equation}
                    \mathcal{C}_{DualFFN} \approx \underbrace{24 N_{img} d^2}_{\text{Image FFN}} + \underbrace{24 N_{prompt} d^2}_{\text{Text FFN}} + \underbrace{8 (N_{img} + N_{prompt}) d^2}_{\text{Q,K,V,O Projections}} = 32 (N_{img} + N_{prompt}) d^2
                \end{equation}
        \end{itemize}
        
    \textit{Note: Standard Gated MLP adds slight overhead, approximated here as standard $d \to xd \to d$ cost.}
\end{itemize}

\textbf{2. Single Stream Block.}
This block concatenates inputs ($N_{total} = N_{img} + N_{prompt}$) and employs a parallel Attention-MLP architecture with a fused output.
\begin{itemize}
    \item \textbf{Single Attention Core}:
    \begin{equation}
        \mathcal{C}_{SingleAttn} \approx 4 (N_{img} + N_{prompt})^2 d
    \end{equation}
    \item \textbf{Single Linear Operations}: The block computes Q,K,V ($3d^2$), an MLP up-projection ($4d^2$), and a large fused output projection that maps the concatenated Attention+MLP output ($d+4d$) back to $d$ ($5d^2$).
    \begin{equation}
        \mathcal{C}_{SingleLinear} \approx \underbrace{(6 + 8 + 10) N_{total} d^2}_{\text{QKV + MLP\_Up + Fused\_Out}} = 24 (N_{img} + N_{prompt}) d^2
    \end{equation}
\end{itemize}

Eventually, we can see that the theoretical complexity is the same for both categories.

\subsection{\toolName{} Complexity.}

\textbf{\toolName{} Complexity.} {\toolName} prunes the reference tokens in the visual stream ($N_{img} \rightarrow \hat{N}_{img}$). This provides a quadratic reduction in the Attention mechanism ($\hat{N}_{img} + N_{prompt})^2$ and a linear reduction in the Visual FFN. The Text FFN cost remains constant but is negligible compared to the visual computation.

\textbf{Overhead Analysis.} The additional cost introduced by \toolName{} involves the Importance Scoring and Selection phases.
\begin{itemize}
    \item \textbf{Scoring:} The computation of Token Influence Score $I_j$ requires aggregating attention maps from a subset of representative layers $|\mathcal{L}|$. This incurs $O(|\mathcal{L}| \cdot |\mathcal{G}| \cdot |\mathcal{C}|)$. However, since this is only performed at sparse update intervals $\Delta T$ (e.g., every 10 steps), the amortized cost is negligible: $\frac{1}{\Delta T} \cdot O(|\mathcal{L}| \cdot |\mathcal{G}| \cdot |\mathcal{C}|) \approx 0$.
    \item \textbf{Selection:} The sorting and gathering operation has a complexity of $O(|\mathcal{C}| \log |\mathcal{C}|)$, which is computationally trivial compared to the matrix multiplications in attention layers.
\end{itemize}

\subsection{Single-Layer FLOPs Breakdown} \label{app:single_layer_flops}

We provide a granular breakdown of the FLOPs for a single transformer layer. We analyze a heavy in-context generation scenario where the visual input consists of noisy tokens ($|\mathcal{G}|=4096$) and dense reference tokens ($|\mathcal{C}|=4096$), totaling $N_{img}=8192$ visual tokens. The condition (text or multimodal prompts)  is $N_{prompt}=512$.

\toolName{} prunes the reference tokens by 50\% ($4096 \rightarrow 2048$), reducing the total visual tokens to $N_{img}'=6144$.

\begin{table}[]
    \centering
    \caption{\textbf{Single-Layer FLOPs Breakdown (GFLOPs).} Comparison under heavy context load ($N_{total} \approx 8.7k \rightarrow 6.6k$). \toolName{} effectively reduces the quadratic attention bottleneck and the massive linear operations associated with the visual stream.}
    \label{tab:layer_breakdown_corrected}
    \vspace{2mm}
    \resizebox{1.0\linewidth}{!}{
    \begin{tabular}{llccc|c}
        \toprule
        \textbf{Architecture} & \textbf{Component} & \textbf{Baseline (GFLOPs)} & \textbf{\toolName{} (GFLOPs)} & \textbf{Reduction (\%)} & \textbf{Mechanism} \\
        \midrule
        \multirow{4}{*}{\shortstack[l]{\textbf{Flux Joint}\\\textit{(Double Stream)}\\$d=4096$}} 
        & Joint Attn Core ($N_{tot}^2$) & 1241.25 & 725.85 & -41.5\% & Quadratic \\
        & Image Linear Ops ($N_{img}$) & 3298.53 & 2473.90 & -25.0\% & Linear \\
        & Text Linear Ops ($N_{prompt}$) & 206.16 & 206.16 & 0.0\% & Fixed Overhead \\
        \cmidrule(lr){2-6}
        & \textbf{Layer Total} & \textbf{4745.94} & \textbf{3405.91} & \textbf{-28.2\%} & \\
        \midrule
        \multirow{3}{*}{\shortstack[l]{\textbf{Flux Single}\\\textit{(Single Stream)}\\$d=4096$}} 
        & Single Attn Core ($N_{tot}^2$) & 1241.25 & 725.85 & -41.5\% & Quadratic \\
        & Single Linear Ops ($N_{tot}$) &  3504.69  & 2680.06 & -23.5\% & Linear (Fused) \\
        \cmidrule(lr){2-6}
        & \textbf{Layer Total} & \textbf{4745.94} & \textbf{3405.91} & \textbf{-28.2\%} & \\
        \midrule
        \multirow{4}{*}{\shortstack[l]{\textbf{Qwen-Image-Edit}\\\textit{(Double Stream)}\\$d=4096$}} 
        & Joint Attn Core ($N_{tot}^2$) & 1241.25 & 725.85 & -41.5\% & Quadratic \\
        & Image Linear Ops ($N_{img}$) & 4398.05 & 3298.53 & -25.0\% & Linear \\
        & Text Linear Ops ($N_{prompt}$) & 274.88 & 274.88 & 0.0\% & Fixed Overhead \\
        \cmidrule(lr){2-6}
        & \textbf{Layer Total} & \textbf{5914.17} & \textbf{4299.26} & \textbf{-27.3\%} & \\
        \bottomrule
    \end{tabular}
    }
\end{table}

\textbf{Observation.}
As shown on Tab.~\ref{tab:layer_breakdown_corrected}, in this high-load regime ($N \approx 8k$), the \textbf{Joint Attention Core} consumes a significant portion of compute ($\approx 30\%$ of total FLOPs).

\subsection{Overhead Breakdown}
\begin{table}[h]
    \centering
    \caption{\textbf{Runtime breakdown analysis of Flux.1-Kontext.} The data represents a typical inference session with $1024 \times 1024$ resolution and 40 sampling steps. With a pruning threshold of 0.85 and an update interval of $\Delta T=10$, the proposed modules add less than 1\% overhead.}
    \label{tab:overhead}
    \begin{tabular}{l c c}
        \toprule
        \textbf{Component} & \textbf{Runtime (s)} & \textbf{Ratio (\%)} \\
        \midrule
        Base Model Inference & 27.692              & 99.45\%             \\
        Importance Scoring   & 0.147               & 0.50\%              \\
        Selection            & 0.014               & 0.05\%              \\
        \midrule
        Total       & 27.853     & 100.0\%    \\
        \bottomrule
    \end{tabular}
\end{table}

\paragraph{Overhead Analysis.}
The additional computational cost introduced by \toolName{} stems primarily from two phases: \textit{Importance Scoring} and \textit{Selection}. We analyze the overhead from both theoretical complexity and empirical runtime perspectives.
To quantify the actual latency of Flux.1-Kontext, we present a representative case in Tab.~\ref{tab:overhead}. The evaluation is conducted under the following configuration: image resolution of $1024 \times 1024$ for both generation and reference, a pruning threshold of 0.85, an update interval $\Delta T=10$, and a total of 40 inference steps.
Under these settings, the \textit{Importance Scoring} consumes only 0.147s (0.50\% of total runtime), and the \textit{Selection} takes a negligible 0.014s (0.05\%). This confirms that \toolName{} introduces virtually no perceptible latency to the standard inference pipeline.

\end{document}